%% file: main.tex
\documentclass{article}
\usepackage{iclr2027_conference,times}

\input{math_commands.tex}

\usepackage{booktabs}
\usepackage{array}
\usepackage{colortbl,xcolor}
\usepackage{amsmath,amssymb,amsthm}
\usepackage{centernot}
\usepackage{graphicx}
\usepackage{microtype}
\usepackage{float}
\usepackage{tikz}
\usetikzlibrary{arrows.meta,positioning,calc}
\usepackage{url}
\usepackage{hyperref}
\usepackage{xcolor}
\usepackage{amsmath}
\usepackage{amssymb}       
\usepackage{graphicx}
\usepackage{array}
\usepackage{tabularx}
\usepackage{wrapfig}
\usepackage{multirow}
\usepackage{booktabs}
\usepackage{multirow}
\usepackage{tabularx}
\usepackage{makecell}
\usepackage[table]{xcolor}
\usepackage{amssymb}

\newcommand{\best}[1]{\cellcolor{blue!9}\textbf{#1}}

\iclrfinalcopy

\definecolor{licensed}{HTML}{2E7D32}
\definecolor{unsupported}{HTML}{B3261E}
\definecolor{softblue}{HTML}{EAF2F8}
\definecolor{softgreen}{HTML}{EAF5EA}
\definecolor{softred}{HTML}{FBEDEC}
\definecolor{softgray}{HTML}{F3F4F6}

\newtheorem{lemma}{Lemma}
\newtheorem{proposition}{Proposition}

\newcommand{\ProjectName}{\textsc{DerivAudit}}

\newcommand{\Evidence}{\mathcal{E}}

\title{Memory Is a Derivation:\\ The Distributed-Evidence Paradox in Long-Term Agents}

\author{%
 \textbf{Hongjun Liu}$^{1}$,
 \textbf{Chen Zhao}$^{1}$
   \vspace{.5em} 
  \\
  $^1$New York University  \\
}

\begin{document}

\maketitle
\lhead{Preprint.}

\begin{abstract}
Long-running LLM agents compress past interactions into persistent memories that may be reused as premises for later tasks.
This creates a distinct \emph{derivation problem}: whether the memory actually follows from what the interaction history supports.
Relevant evidence may be scattered across earlier interactions, while compression can introduce relations or event status that the history never established.
A valid memory may therefore appear unsupported because its citations omit relevant evidence, while individually supported facts may be composed into a stronger statement the history never established.
We characterize this problem through three coupled requirements:
\emph{(1) Evidence scope};
\emph{(2) Compositional validity};
\emph{(3) Admission reliability}.
We therefore ask whether the interaction history available at write time supports what enters persistent memory.
We introduce \textbf{\ProjectName{}}, a framework for auditing whether a memory is actually supported by the history available when it was written.
The audit separates three questions: whether supporting evidence lies beyond writer-provided citations, whether the composed memory introduces unsupported meaning, and how write-time admission decisions affect later memory use.
Across two natural memory corpora, audits using broader pre-write history recover support for nearly 60\% of memories that appear unsupported from citations alone, while 17--21\% remain unsupported after expansion.
Yet broader evidence does not by itself make admission reliable: unsupported memories are still frequently admitted (59--89\%) across verification models, and evidence expansion alone worsens it on two backbones.
Controlled experiments reveal a \emph{distributed-evidence paradox}: valid memories often require combining evidence across interactions, yet several plausible facts can make unsupported compositions harder to detect.
Moreover, requiring a verifier to check more independently true details can itself increase rejection of valid memories.
On 391 labelled real memory writes, composition-aware verification with expanded history improves retention of valid memories rescued by broader-history evidence but does not consistently reduce unsupported admission across backbones.
Downstream probes further show that both distorted and missing memories can harm later answers.
\end{abstract}

\input{paper/01_introduction}

\input{paper/02_related_work}

\input{paper/03_method}

\input{paper/04_experiments}

\input{paper/05_conclusion}

\subsection*{Reproducibility Statement}

Sections~\ref{sec:method} and~\ref{sec:experiments} describe the memory-writing setting, evidence scopes, reference-label procedure, verification views, admission decision rules, evaluation metrics, and controlled interventions used in our analyses.
The appendix provides annotation attrition and label denominators, robustness and protocol controls, threshold-selection details, and additional implementation information.
The accompanying artifact records model configurations, prompts and chat templates, retrieval indexes and evidence budgets, random seeds and sampled record IDs, annotation and adjudication outputs, decision thresholds, experiment manifests, and analysis code.
Results reused across analyses are linked to their original recorded runs rather than reconstructed from prose.

\subsection*{AI Use Statement}

Generative AI tools were used to assist with code and LaTeX polishing.
Separate model instances also served as annotators and blinded adjudicators for the natural memory audit; agreement from this procedure is reported as model-instance agreement rather than human validation.
Human review was conducted separately on a stratified sample, as described in the appendix.
The authors remain responsible for citations, mathematical claims, data processing, and reported results.

\subsection*{Ethics Statement}

Our experiments use public long-term conversation benchmarks and do not collect new private user conversations.
Human annotators were used only to review sampled benchmark records for validation of the model-produced reference labels and were compensated hourly at standard rates.
No new personal profiles or inferred real-world identities were created as part of the study.
Released artifacts should preserve the licenses and redistribution requirements of the underlying datasets, minimize unnecessary reproduction of conversation text, and avoid adding identities or personal attributes not present in the source data.

\bibliography{iclr2027_conference}
\bibliographystyle{iclr2027_conference}

\input{paper/appendix}

\end{document}

%% file: math_commands.tex
\usepackage{amsmath,amsfonts,bm}

\def\eqref#1{equation~\ref{#1}}

\def\1{\bm{1}}

\DeclareMathAlphabet{\mathsfit}{\encodingdefault}{\sfdefault}{m}{sl}
\SetMathAlphabet{\mathsfit}{bold}{\encodingdefault}{\sfdefault}{bx}{n}



%% file: paper/01_introduction.tex
\section{Introduction}
\label{sec:introduction}

Long-running LLM agents use persistent memory to carry information across turns, sessions, and tasks without retaining every past interaction in context.
Prior systems store conversational experience, reflections, and reusable knowledge in persistent state, and increasingly organize or evolve that state over time~\citep{park2023generativeagentsinteractivesimulacra,zhong2023memorybankenhancinglargelanguage,shinn2023reflexionlanguageagentsverbal,zhao2024expelllmagentsexperiential,xu2025amemagenticmemoryllm}.
Recent surveys accordingly treat memory formation, evolution, and retrieval as core components of long-running agents~\citep{zhang2024surveymemorymechanismlarge}.
This makes reliable memory writing important: once information is stored, later tasks may treat it as an established fact rather than return to the original interaction history, and may further combine stored memories to draw new conclusions.
Yet a memory may not faithfully reflect what the history actually supports.
A writer can combine observations from different turns, omit an important qualification, or introduce a relation that was never established.
When such an error enters persistent memory, it can continue to shape later agent behavior even after the original interaction is no longer in context.

This raises question for reliable memory writing: how can we tell whether a new memory is actually supported by the interaction history?
Figure~\ref{fig:overview} illustrates why this is not straightforward.
In one case, a memory is supported by earlier interactions, but its attached citations capture only part of the relevant evidence, so a verifier that checks citations alone may reject a valid memory.
In the other, individual facts are each supported somewhere in the history, but the memory combines them into a relation or changes event's status in a way the history never established.
A reliable verifier must look beyond citation coverage and determine whether the memory as a whole is justified by the history~\citep{rashkin-etal-2023-measuring,gao-etal-2023-enabling,gao-etal-2023-rarr}.
We study this as a \emph{derivation problem}: whether proposed memory can be derived from the interactions that came before it.
This problem has three closely related aspects.
\textbf{(1) Evidence scope:} the evidence needed to support a memory may extend beyond the sources explicitly attached to it~\citep{joshi2026eywa,jin2026memir}.
\textbf{(2) Compositional validity:} even when individual facts are supported, their combination may introduce meaning that the history does not support.
\textbf{(3) Admission reliability:} a write-time verifier should reject unsupported memories without discarding valid information that may be needed later~\citep{cui2026memtxn,zhang2026consistencygate}.

We introduce \textbf{\ProjectName{}}, a framework for auditing how interaction history is turned into persistent memory.
We call the underlying correctness requirement \emph{semantic derivation integrity}: what is written into persistent memory should be supported by the interaction history available at write time.
To audit this requirement, \ProjectName{} compares writer-provided citations with broader pre-write history, checks whether the full meaning of a memory is supported rather than only its individual facts, and evaluates which supported and unsupported memories survive write-time verification.

\input{figures_tex/figure1}

We evaluate \ProjectName{} on 400 unedited memory writes from LoCoMo~\citep{maharana-etal-2024-evaluating} and HaluMem-Medium~\citep{chen2026halumemevaluatinghallucinationsmemory}, drawn from two memory-writing pipelines and evaluated with three verification models.
Across both corpora, broader pre-write history recovers support for about 60\% of memories whose attached citations are insufficient, showing that citation-only verification can mistake incomplete provenance for unsupported memory; even after expansion, roughly one fifth of writes remain unsupported.
Controlled experiments reveal a complementary difficulty: valid memories often require evidence distributed across multiple interactions, yet several individually plausible facts can make unsupported combinations harder to reject, a pattern we call the \emph{distributed-evidence paradox}.
No verification approach consistently resolves this tension across models, and requiring the verifier to check more independently supported details can itself increase rejection of valid memories.
On 391 labelled natural memory writes, broader-history and composition-aware verification improve retention of valid memories rescued by additional historical evidence on some models, while rejection of unsupported memories remains inconsistent.
Finally, downstream interventions show why these write-time errors matter: distorted memories can propagate incorrect information, while rejecting valid memories can remove information needed for later tasks.

Our contributions are threefold.
\textbf{(1) A derivation view of memory writing.}
We frame persistent memory as a derivation from interaction history and define \emph{semantic derivation integrity}: what is written should be supported by the history available at write time.
\textbf{(2) An empirical framework for studying write-time memory reliability.}
\ProjectName{} provides a common audit setup for separating incomplete provenance from unsupported composition and examining admission decisions across verification settings.
\textbf{(3) A systematic study of memory-writing failure modes.}
Across natural and controlled settings, we characterize incomplete provenance, the \emph{distributed-evidence paradox}, and verification-induced rejection of valid memories, and show that both distorted and missing memories can affect downstream behavior.

%% file: figures_tex/figure1.tex
\begin{figure}[!t]
\centering
\includegraphics[width=\linewidth]{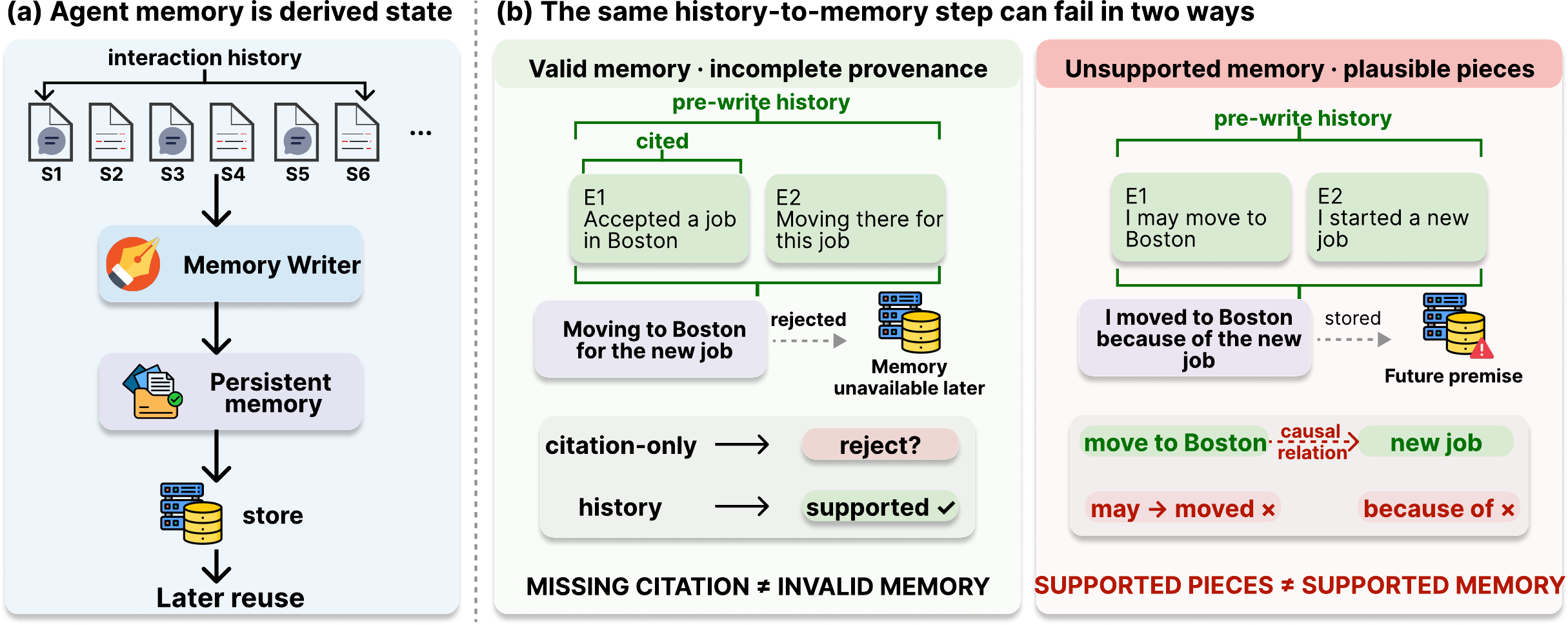}
\vspace{-20pt}
\caption{
\textbf{Persistent memory as derived state.}
\textbf{(a)} Interaction history is compressed into persistent memory that can later be reused as agent state.
\textbf{(b)} This write step has two distinct failure modes: incomplete citations can make a valid memory appear unsupported, while supported historical pieces can be combined into a memory whose full meaning is not supported. 
}
\label{fig:overview}
\vspace{-10pt}
\end{figure}

%% file: paper/02_related_work.tex
\section{Related Work}
\label{sec:related}

We situate our work along three lines. \textbf{(1) Long-term memory for LLM agents.} Persistent-memory systems enable agents to retain and reuse information across extended interactions through extraction, consolidation, updating, and retrieval \citep{packer2023memgpt,chhikara2025mem0,latimer2026hindsight}, while benchmarks evaluate recall, temporal reasoning, knowledge updating, and failures across memory operations \citep{latimer2026hindsight,wu2025longmemeval,hu2025memoryagentbench,chen2026halumemevaluatinghallucinationsmemory}. Rather than asking primarily whether stored information can later be recalled, we study whether a memory was semantically justified by the interaction history when it entered persistent state. \textbf{(2) Reliable memory writing and admission.} Recent work increasingly treats memory state as a reliability boundary: Eywa links memories to provenance, MemIR distinguishes source evidence from truth-bearing memory content, and ConsistencyGate and MemTxn control whether candidate updates should be committed \citep{joshi2026eywa,jin2026memir,zhang2026consistencygate,cui2026memtxn}. Our setting exposes two additional challenges: writer-supplied provenance may omit historical evidence that supports a valid memory, while individually supported facts may still fail to license the relations introduced when they are composed into a memory. \ProjectName{} therefore evaluates natural memory writes under controlled evidence scopes and measures both unsupported memories retained and historically supported memories discarded. \textbf{(3) Fine-grained factuality and compositional verification.} Prior work uses dependency-level entailment, semantic graphs, predicate--argument structure, and turn-level verification to expose unsupported relations \citep{goyal2020dae,laban-etal-2022-summac, honovich-etal-2022-true-evaluating, ribeiro2022factgraph,cattan2026qasem,lewis2025vista,liu2025suceareasoningintensiveretrievaladversarial}, while studies of abstractive generation document cases in which source-present content is recombined into unsupported statements \citep{maynez2020faithfulness,pagnoni2021frank}. We extend these ideas to persistent agent memory, where a record is derived from longitudinal interaction history and may later be reused without its original sources; \ProjectName{} jointly studies \emph{where} its support resides, \emph{what} meaning composition adds, and \emph{which} memories survive write-time verification and affect later agent behavior. Table~\ref{tab:related-comparison} summarizes these distinctions.

\input{Tables/relatedwork}

%% file: Tables/relatedwork.tex
\begin{table*}[ht]
\centering
\vspace{-4mm}
\caption{
Comparison with closely related work.
\ProjectName{} studies whether interaction history supports what is written into
persistent memory, including evidence beyond attached provenance, compositional
meaning, and the consequences of admission decisions.
}
\label{tab:related-comparison}
\resizebox{\textwidth}{!}{%
\begin{tabular}{@{}lccccc@{}}
\toprule
\textbf{Work} &
\shortstack{\textbf{Agent}\\\textbf{Memory}} &
\shortstack{\textbf{Evidence Beyond}\\\textbf{Attached Provenance}} &
\shortstack{\textbf{Compositional}\\\textbf{Semantics}} &
\shortstack{\textbf{Write-time}\\\textbf{Admission}} &
\shortstack{\textbf{Downstream}\\\textbf{Memory Effects}} \\
\midrule

\multicolumn{6}{@{}l}{\textit{Long-term memory and reliability}} \\

LoCoMo / LongMemEval\textsuperscript{a}
& $\checkmark$ & -- & -- & -- & -- \\

HaluMem~\citep{chen2026halumemevaluatinghallucinationsmemory}
& $\checkmark$ & -- & -- & -- & $\checkmark$ \\

Eywa~\citep{joshi2026eywa}
& $\checkmark$ & -- & -- & $\checkmark$ & -- \\

ConsistencyGate~\citep{zhang2026consistencygate}
& $\checkmark$ & -- & -- & $\checkmark$ & -- \\

MemTxn~\citep{cui2026memtxn}
& $\checkmark$ & -- & -- & $\checkmark$ & -- \\

MemIR~\citep{jin2026memir}
& $\checkmark$ & -- & -- & -- & -- \\

\midrule
\multicolumn{6}{@{}l}{\textit{Fine-grained factuality}} \\

DAE / FactGraph / QASemConsistency\textsuperscript{b}
& -- & -- & $\checkmark$ & -- & -- \\

VISTA~\citep{lewis2025vista}
& -- & -- & -- & -- & -- \\

\midrule

\textbf{\ProjectName{} (Ours)}
& $\checkmark$
& $\checkmark$
& $\checkmark$
& $\checkmark$
& $\checkmark$ \\

\bottomrule
\end{tabular}%
}

\begin{minipage}{\textwidth}
\scriptsize
\textsuperscript{a}\citet{maharana-etal-2024-evaluating,wu2025longmemeval};
\textsuperscript{b}\citet{goyal2020dae,ribeiro2022factgraph,cattan2026qasem}.
\textbf{Evidence Beyond Attached Provenance} denotes explicit comparison or
recovery of support outside the writer-provided evidence for the same memory write.
\textbf{Compositional Semantics} denotes explicit verification of relations or
qualifications beyond isolated factual ingredients.
\textbf{Downstream Memory Effects} denotes analysis of how stored or missing
memory affects later agent behavior.
\end{minipage}
\end{table*}

%% file: paper/03_method.tex
\section{\ProjectName: Auditing Memory as Derivation}
\label{sec:method}

We use \emph{agent memory} to mean a compact record distilled from earlier interactions and stored for use in later turns or tasks.
A \emph{candidate memory write} is a compact record proposed for persistent storage; admission determines whether that record is committed to memory.
\ProjectName{} audits this process by asking three questions: where the memory's support lies in the earlier interaction history, whether that history supports the full meaning of the memory, and whether write-time verification retains the right information for later use.

\subsection{Evidence Scope: Recovering the History Behind a Memory}
\label{sec:evidence-expansion}

Writer-provided citations may capture only part of the history that supports a memory.
Our goal is therefore to distinguish a genuinely unsupported memory from a valid memory whose provenance is incomplete.
For each candidate memory, we compare two views of the same pre-write history.
The \emph{citation-only} view contains only the passages attached by the writer.
The \emph{expanded-history} view supplements these citations with up to 12 BM25-retrieved passages from interactions that occurred before the memory was written, subject to a total budget of 16 passages.
Later interactions are excluded, and previously written memories are not treated as independent evidence for claims derived from the same history.

We evaluate the unchanged memory independently under both evidence views using the same reference-labeling procedure described in Section~\ref{sec:obligations}.
Passages are presented chronologically and without indicating whether they were cited or retrieved, so the paired judgments isolate the effect of seeing more of the pre-write history.
We call a memory \emph{provenance-repaired} when it is insufficiently supported by its citations but supported after history expansion, and report the fraction of citation-insufficient memories repaired in this way.
Because history expansion is retrieval-bounded, an insufficient judgment does not always establish that a memory was unsupported by the full history.
We therefore treat search-limited cases as unresolved rather than as historically unsupported.

\subsection{Compositional Validity: Checking What the Agent Will Remember}
\label{sec:obligations}
\label{sec:distributed}

Recovering the relevant history does not guarantee that a memory is supported.
A writer may combine individually supported facts into a stronger statement, for example by adding a causal relation, changing when an event occurred, or turning a plan into an accomplished fact.
We therefore ask whether the evidence supports the \emph{full meaning} that will be stored in memory.

\noindent\textbf{(1) Semantic obligations.}
We decompose each memory into the factual claims, relations, and qualifications that must all be supported for the memory as a whole to be valid.
We refer to these requirements as \emph{semantic obligations}.
For memory \(m_i\), let
\[
\mathcal{O}(m_i)=\{o_{i1},\ldots,o_{iJ_i}\}.
\]
For example, ``The user moved to Boston because of the new job'' requires evidence not only for the move and the job, but also that both occurred and that the job caused the move.
We consider a memory supported only when all of its obligations are supported by the available evidence:
\[
Y(m_i,E_i)
=
\mathbf{1}\!\left[
\forall o\in\mathcal{O}(m_i),\;
\ell(o,E_i)=\textsc{Supported}
\right].
\]

\noindent\textbf{(2) Reference labels.}
Each obligation receives a four-way reference label from two independent model assessments, with disagreements resolved by blinded adjudication.
Support and contradiction are assessed separately so that missing evidence is not treated as evidence against a claim.

\noindent\textbf{(3) Verification views.}
We next study how different representations of the same memory affect write-time verification.
Every view receives the same candidate memory, evidence, and verification model; only the representation used for checking changes:
\[
s_i^{(v)}
=
V_\theta\!\left(\phi_v(m_i),E_i\right),
\]
where \(\phi_v\) denotes verification view \(v\).
We compare holistic factuality, atomic-claim verification, and predicate--argument QA~\citep{luo2023chatgpt,min2023factscore,cattan2026qasem} with views that explicitly represent relations and qualifications beyond isolated claims.
This comparison tests whether different verification representations preserve the compositional meaning needed to determine whether the memory is supported by its history.

\noindent\textbf{(4) Controlled evidence distribution.}
Finally, we isolate the effect of where supporting information appears in the history.
Across 70 matched families of examples, we construct valid and invalid memories under \emph{local} and \emph{distributed} evidence conditions while keeping the candidate memory and its validity fixed.
Valid distributed cases require combining evidence across passages; invalid cases contain individually plausible premises but no support for the relation that combines them.
Additional controls remove complete sets of evidence needed to support the target relation or replace plausible premises with matched irrelevant passages.
These interventions distinguish difficulty integrating distributed evidence from difficulty rejecting unsupported compositions.

\input{figures_tex/pipeline}

\subsection{Admission Reliability: Retaining Knowledge for Later Tasks}
\label{sec:lifecycle}

A write-time verifier does more than classify a memory: its decision determines what information remains available to the agent later.
Admitting an unsupported memory can introduce an incorrect premise, while rejecting a supported one can remove useful information.
We therefore study whether verification makes reliable \emph{admission decisions} and what happens when those decisions are wrong.

\noindent\textbf{(1) Write-time admission.}
We replay the same natural candidate writes under three settings: citation-only holistic verification, expanded-history holistic verification, and expanded-history obligation-aware verification.
For verification setting \(v\), a memory is admitted when its score exceeds the corresponding threshold:
\[
a_i^{(v)}
=
\mathbf{1}\!\left[s_i^{(v)}\geq\tau_v\right].
\]
The two expanded-history settings receive the same evidence, allowing us to separate the effect of seeing more history from the effect of checking the memory differently.
We measure both sides of the decision: retention of supported memories and admission of unsupported ones, with separate analysis of provenance-repaired memories.

\noindent\textbf{(2) Verification granularity.}
We next ask whether checking a memory in greater detail can itself make valid memories harder to retain.
For supported memories, we ask the verifier to check two or four additional obligations that are independently known to be true, while keeping the candidate memory, evidence, and reference validity unchanged.
The additional obligations change what the verifier must check, not the content of the memory.
Any systematic decrease in verification score or admission therefore arises from requiring more correct checks rather than from introducing an actual defect.
We call this effect \emph{accumulated verification noise}.

\noindent\textbf{(3) Later memory use.}
We examine the consequences after a write-time decision is made.
We track whether later rewriting preserves qualifications such as timing and uncertainty, and whether stored memories are retrieved into later contexts.
In controlled downstream probes, we fix the task context and provide either the correct memory, a distorted version, or no memory.
This isolates the cost of carrying an incorrect premise forward from the cost of removing information that a later task needs.

%% file: figures_tex/pipeline.tex
\begin{figure}[!t]
\centering
\includegraphics[width=\linewidth]{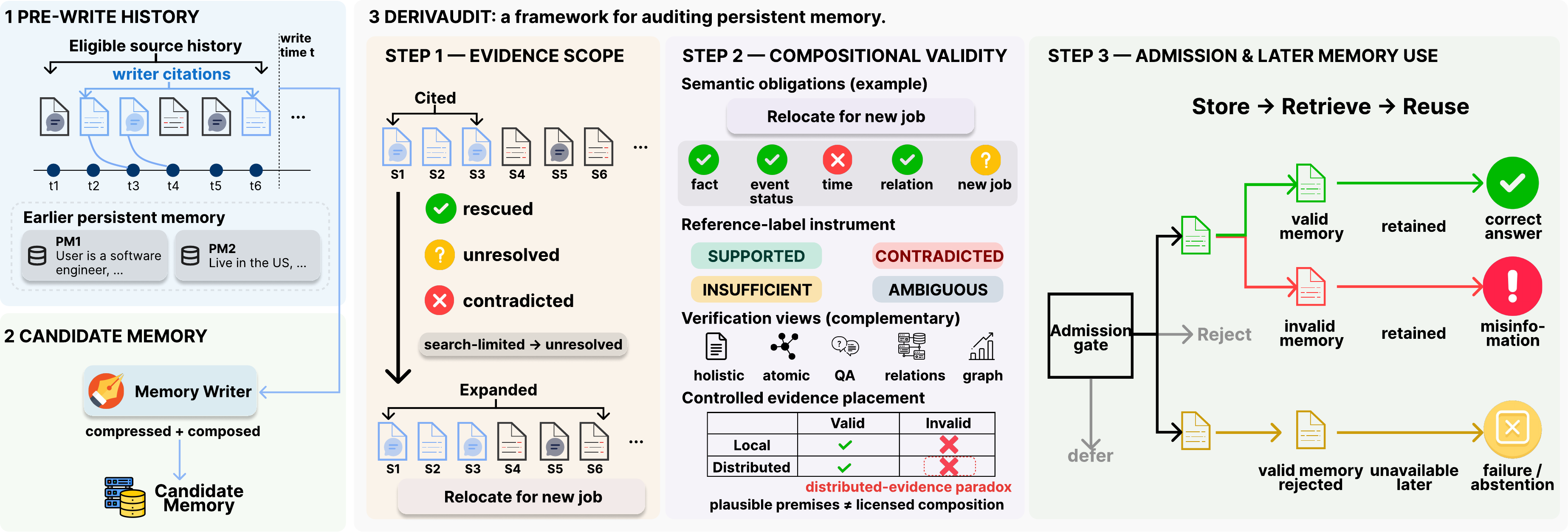}
\vspace{-20pt}
\caption{
\textbf{\ProjectName{} audits memory derivation from pre-write history to later use.}
The framework organizes the analysis around three questions: where a memory's historical support resides, whether that support licenses the full meaning introduced during composition, and how alternative admission decisions affect the memory available to later tasks.
}
\vspace{-15pt}
\label{fig:pipeline}
\end{figure}

%% file: paper/04_experiments.tex
\section{Experiments}
\label{sec:experiments}

We use \ProjectName{} to study three aspects of memory writing: evidence scope, compositional validity, and admission reliability.
We combine natural memory writes with controlled interventions and then examine how write-time errors affect later memory use.

\textbf{Evaluation Data.}
Our natural-data audit contains 400 unedited memory writes, split evenly between LoCoMo~\citep{maharana-etal-2024-evaluating} and HaluMem-Medium~\citep{chen2026halumemevaluatinghallucinationsmemory}.
Each write is evaluated under citation-only and expanded pre-write evidence; 391 receive resolved expanded-history reference labels and enter the admission evaluation.
Our controlled evaluations include 70 matched families (280 examples) that vary whether evidence is local or distributed while holding the candidate memory and its validity fixed, and a 512-record suite for comparing verification approaches on single- and multi-passage support and unsupported relations.
Additional experiments examine repeated memory rewriting, retrieval, and downstream use.

\textbf{Models and Verification Settings.}
We evaluate Qwen3-30B-A3B, Gemma-4-31B, and Llama-3.3-70B-Instruct.
Across experiments, these models serve as verification models and, where applicable, as memory writers or downstream readers.
For write-time verification, we compare established approaches, holistic factuality judgment~\citep{zheng2023judging}, atomic-claim verification~\citep{min2023factscore}, and predicate--argument QA~\citep{cattan2026qasem}, with representations designed to make compositional relations explicit, including compact-relation, composition-graph, and obligation-level views.
Within each comparison, all views receive the same candidate memories and evidence.

\textbf{Metrics and Labels.}
Following Sections~\ref{sec:evidence-expansion}--\ref{sec:lifecycle}, we measure provenance repair for evidence scope; detection of unsupported relations and retention of valid multi-passage memories for compositional validity; and supported-memory retention and unsupported-memory admission for admission reliability.
Natural reference labels are derived from two independent model assessments, with disagreements resolved by blinded adjudication.
Threshold selection and model-specific decision rules appear in Appendix~\ref{app:thresholds}.

\subsection{What Does Write-Time Verification Retain and Reject?}
\label{sec:overall-performance}

We first examine write-time admission on the 391 natural memories with resolved expanded-history labels: 313 are supported and 78 are unsupported or contradicted.
Table~\ref{tab:overall-main} compares three settings: citation-only holistic verification, expanded-history holistic verification, and expanded-history obligation-aware verification.

\textbf{Broader History Recovers Valid Memories with Incomplete Citations.}
On Qwen, retention of provenance-repaired memories rises from 82.1\% with citation-only verification to 93.8\% with expanded history, an 11.6-point paired improvement ($p=0.007$).
Adding obligation-aware checking yields similar retention at 92.9\% ($p=0.023$ versus citation-only).
Overall valid-memory retention remains nearly unchanged across the three settings at 90.4\%, 90.1\%, and 90.7\%.
The same pattern appears on the other verification models: expanded history raises provenance-repaired retention from 58.0\% to 95.5\% on Gemma and from 85.7\% to 98.2\% on Llama.

\textbf{Unsupported Writes Remain Frequently Admitted.}
On Qwen, unsupported-memory admission decreases from 84.6\% with citation-only verification to 80.8\% with expanded history and 76.9\% with obligation-aware checking, but the latter difference from citation-only verification is not statistically resolved on the 78 negative records ($p=0.24$).
Across models, obligation-aware checking is not consistently associated with lower unsupported admission: the rate falls from 66.7\% to 59.0\% on Gemma but rises from 83.3\% to 88.5\% on Llama.
Thus, broader history consistently improves retention of valid memories with incomplete provenance across the three models, whereas rejection of unsupported writes remains much less stable.

\input{Tables/main_table}

\input{Tables/compositional_verify}

\input{figures_tex/figure3}

\subsection{Why Is Memory Admission Difficult?}
\label{sec:analysis}

Memory admission determines whether a candidate write is stored for later use.
The preceding results reveal an asymmetry: broader history often helps retain supported memories with incomplete citations, while unsupported writes remain frequently admitted.
We examine this gap through evidence scope, compositional validity, and admission reliability.

\input{figures_tex/figure5}

\textbf{Evidence Scope: Valid Memories Often Need Evidence Beyond Their Citations.}
Figure~\ref{fig:figure3}(a) compares citation-only and expanded-history reference judgments on the same natural writes.
Citations alone are insufficient for 42.5\% of LoCoMo and 51.0\% of HaluMem-Medium memories, but broader history recovers support for 60.0\% and 59.8\% of these cases, raising the overall supported fraction from 53.5\% to 79.5\% and from 47.0\% to 77.0\%, respectively.
Moreover, 25 of 107 citation-supported LoCoMo memories and 14 of 94 HaluMem-Medium memories require multiple passages jointly for support.
Thus, expanding the evidence recovers incomplete provenance, but also raises a harder question: whether evidence distributed across interactions supports the memory as a whole.

\textbf{Compositional Validity: Supported Pieces Do Not Guarantee a Supported Memory.}
Table~\ref{tab:semantic-main} shows a trade-off between detecting unsupported relations and retaining valid multi-passage memories.
On Qwen, atomic verification detects only 16.5\% of unsupported relations, while predicate--argument QA reaches 39.2\% but retains only 56.4\% of valid multi-passage memories.
Our compact-relation view retains 94.9\% while detecting 59.5\%, close to the 60.8\% detection rate of holistic verification.
No single verification view consistently performs well on both criteria across models.
Figure~\ref{fig:figure3}(b--c) further isolates the role of distributed evidence while holding the candidate memory and its validity fixed.
Under distributed evidence, holistic and obligation-aware verification detect only 8.6\% and 11.4\% of invalid Qwen cases, respectively.
We call this the \emph{distributed-evidence paradox}: valid memories often require combining evidence across interactions, yet the same setting can make unsupported combinations harder to reject.
Matched controls suggest that this pattern is not explained solely by longer context or greater evidence distance.
Removing all sufficient evidence for the target relation flips 88--100\% of affected decisions to rejection, while replacing plausible premises with matched irrelevant history reduces invalid-memory acceptance from 65--88\% to 0--10\%.

\textbf{Admission Reliability: More Detailed Checking Can Reject Valid Memories.}
More detailed verification introduces a complementary failure mode.
When the verifier is asked to check two or four additional obligations that are independently known to be true, while the memory itself remains unchanged, its confidence often declines.
With four additional checks, Qwen's median support score drops by 7.7 log-odds units, with 98.7\% of records decreasing.
The resulting admission rate changes little on Qwen, from 99.4\% to 98.1\%, but falls from 93.3\% to 2.6\% on Gemma and from 97.4\% to 35.5\% on Llama.
We refer to this pattern as \emph{accumulated verification noise}.
Together, these results expose a two-sided tension: insufficient checking can miss unsupported compositions, while increasingly detailed checking can make valid memories harder to retain.
Alternative aggregation rules shift this trade-off but do not eliminate it (Appendix~\ref{app:controls}).

\subsection{What Happens After the Write-Time Decision?}
\label{sec:case-study}

We finally examine what happens after information enters persistent memory: whether timing, uncertainty, and other qualifications remain stable during later rewriting, and how distorted or missing memories affect downstream tasks.

\textbf{Stored Memories Can Drift during Later Rewriting.}
Figure~\ref{fig:figure5}(a) follows naturally generated memories through repeated extraction and consolidation.
Both temporal drift, which changes when an event is said to occur, and modality drift, which changes whether an event is planned, uncertain, or realized, increase substantially across rounds.
For Gemma, the two rates rise from 8.4\% and 15.8\% at extraction to 69.7\% and 78.1\% by round 4; for Qwen, from 7.0\% and 12.0\% to 91.1\% and 92.6\%.
These trends are descriptive rather than causal, since citation breadth, writer settings, and record populations also vary across rounds.

\textbf{Admission Errors Affect Downstream Answers.}
In Figure~\ref{fig:figure5}(b), we vary only the memory supplied to a downstream reader while holding its retrieval position and surrounding context fixed.
Relative to a correct memory, supplying a distorted memory increases downstream failure by 93.0, 83.8, and 92.2 percentage points for Gemma, Qwen, and Llama, while removing a valid memory increases failure by 99.5, 92.8, and 94.9 points on queries that require that information.
Natural retrieval is consistent with the same pattern (Figure~\ref{fig:figure5}(c)): invalid retrieved memories produce misinformation in 100\% and 96.2\% of Gemma and Qwen cases, while valid retrieved memories support correct answers in 91.9\% and 85.9\%.

\textbf{A Natural Write Illustrates a Derivation Failure.}
Figure~\ref{fig:natural-case} shows an unedited HaluMem example.
The history states that Sharon Brown \emph{aspires} to establish a sports academy and separately describes her social network as supportive.
The written memory introduces two unsupported changes: a future plan becomes the present-tense assertion that she \emph{is actively building} the academy, while a generally supportive social network is recast as supporting that specific effort.
Neither stronger statement is established by the pre-write history, and the memory remains unsupported after evidence expansion.

\input{figures_tex/figure6}

%% file: Tables/main_table.tex
\begin{table*}[ht]
\centering
\caption{
Memory admission under three write-time verification settings on 391 decidable
natural writes.
Expanded history adds evidence from pre-write interactions beyond writer-provided
citations; composition-aware checking additionally verifies meaning introduced
when information is composed into memory.
Lower unsupported admission and higher retention are better.
A no-gate policy accepts every write and is omitted.
}
\label{tab:overall-main}

\scriptsize
\setlength{\tabcolsep}{3.0pt}
\renewcommand{\arraystretch}{0.90}

\begin{tabular*}{\textwidth}{
@{\extracolsep{\fill}}
llccccc
@{}
}
\toprule
\textbf{Model}
&
\textbf{Verification Setting}
&
\shortstack{\textbf{History}\\\textbf{Expansion}}
&
\shortstack{\textbf{Composition}\\\textbf{Check}}
&
\shortstack{\textbf{Unsupported}\\\textbf{Admission}$\downarrow$}
&
\shortstack{\textbf{Valid}\\\textbf{Retention}$\uparrow$}
&
\shortstack{\textbf{Repaired}\\\textbf{Retention}$\uparrow$}
\\
\midrule

\multirow{3}{*}{\textbf{Qwen}}
& Citation-only
& -- & --
& 0.846 & 0.904 & 0.821 \\

& + Expanded history
& \checkmark & --
& 0.808 & 0.901 & 0.938 \\

& + Obligation-aware check
& \checkmark & \checkmark
& 0.769 & 0.907 & 0.929 \\

\midrule

\multirow{3}{*}{\textbf{Gemma}$^\dagger$}
& Citation-only
& -- & --
& 0.667 & 0.837 & 0.580 \\

& + Expanded history
& \checkmark & --
& 0.769 & 0.971 & 0.955 \\

& + Obligation-aware check
& \checkmark & \checkmark
& 0.590 & 0.939 & 0.920 \\

\midrule

\multirow{3}{*}{\textbf{Llama}$^\dagger$}
& Citation-only
& -- & --
& 0.833 & 0.946 & 0.857 \\

& + Expanded history
& \checkmark & --
& 0.897 & 0.974 & 0.982 \\

& + Obligation-aware check
& \checkmark & \checkmark
& 0.885 & 0.974 & 0.973 \\

\bottomrule
\end{tabular*}

\vspace{1mm}
\begin{minipage}{\textwidth}
\scriptsize
\textit{Notes.}
Repaired memories are valid writes supported only after expanding beyond writer-provided citations.
Qwen uses cross-fitted thresholds; $^\dagger$Gemma and Llama use argmax operating points.
On Qwen, repaired retention improves with expanded history ($+0.116$, $p=0.007$) and composition-aware checking ($+0.107$, $p=0.023$), while the reduction in unsupported admission is not statistically resolved ($p=0.24$).
\end{minipage}
\vspace{-20pt}
\end{table*}

%% file: Tables/compositional_verify.tex
\begin{table*}[t]
\centering
\caption{
Compositional verification on the 512-record controlled suite.
Established and composition-aware verification views are evaluated on the same
candidate memories and evidence in a shared scoring harness.
$U$ denotes unsupported-relation detection, and Multi denotes retention of valid
memories requiring joint support from multiple passages.
}
\label{tab:semantic-main}

\scriptsize
\setlength{\tabcolsep}{3.2pt}
\renewcommand{\arraystretch}{0.91}

\begin{tabular*}{\textwidth}{
@{\extracolsep{\fill}}
lllccc
@{}
}
\toprule
\textbf{Model}
&
\textbf{View Type}
&
\textbf{Verification View}
&
\shortstack{\textbf{Unsupported}\\\textbf{Detection} $U\uparrow$}
&
\shortstack{\textbf{Multi-span}\\\textbf{Retention}$\uparrow$}
&
\shortstack{\textbf{Valid}\\\textbf{Rejection}$\downarrow$}
\\
\midrule

\multirow{6}{*}{\textbf{Qwen}}
&
\multirow{3}{*}{Established}
& Holistic judge~\citep{luo2023chatgpt}
& \best{0.608} & 0.846 & 0.088 \\

&
& Atomic claims~\citep{min2023factscore}
& 0.165 & 0.744 & 0.121 \\

&
& Predicate--argument QA~\citep{cattan2026qasem}
& 0.392 & 0.564 & 0.115 \\

\cmidrule(lr){2-6}

&
\multirow{3}{*}{\textbf{Composition-aware}}
& \textbf{Compact relations}
& 0.595 & \best{0.949} & \best{0.082} \\

&
& \textbf{Composition graph}
& 0.582 & 0.872 & 0.093 \\

&
& \textbf{Per-obligation}
& 0.291 & 0.897 & 0.093 \\

\midrule

\multirow{6}{*}{\textbf{Gemma}$^\dagger$}
&
\multirow{3}{*}{Established}
& Holistic judge~\citep{luo2023chatgpt}
& 0.937 & 0.923 & 0.022 \\

&
& Atomic claims~\citep{min2023factscore}
& 0.000 & \best{1.000} & \best{0.005} \\

&
& Predicate--argument QA~\citep{cattan2026qasem}
& 0.975 & 0.205 & 0.176 \\

\cmidrule(lr){2-6}

&
\multirow{3}{*}{\textbf{Composition-aware}}
& \textbf{Compact relations}
& \multicolumn{3}{c}{\textit{degenerate: rejects all 39 multi-span memories}} \\

&
& \textbf{Composition graph}
& 0.823 & 0.974 & 0.011 \\

&
& \textbf{Per-obligation}
& \best{0.987} & 0.769 & 0.055 \\

\bottomrule
\end{tabular*}

\vspace{1mm}
\begin{minipage}{\textwidth}
\scriptsize
\textit{Notes.}
Candidate memory and evidence are fixed within each comparison.
Established rows instantiate prior verification paradigms in the same shared harness.
Single-span valid retention is at least 0.91 in every reported cell and is omitted.
Qwen uses out-of-fold tuned thresholds;
$^\dagger$Gemma uses its argmax operating point.
Highlighted cells mark within-model extrema.
\end{minipage}
\vspace{-15pt}
\end{table*}

%% file: figures_tex/figure3.tex
\begin{figure}[t]
\centering
\includegraphics[width=\linewidth]{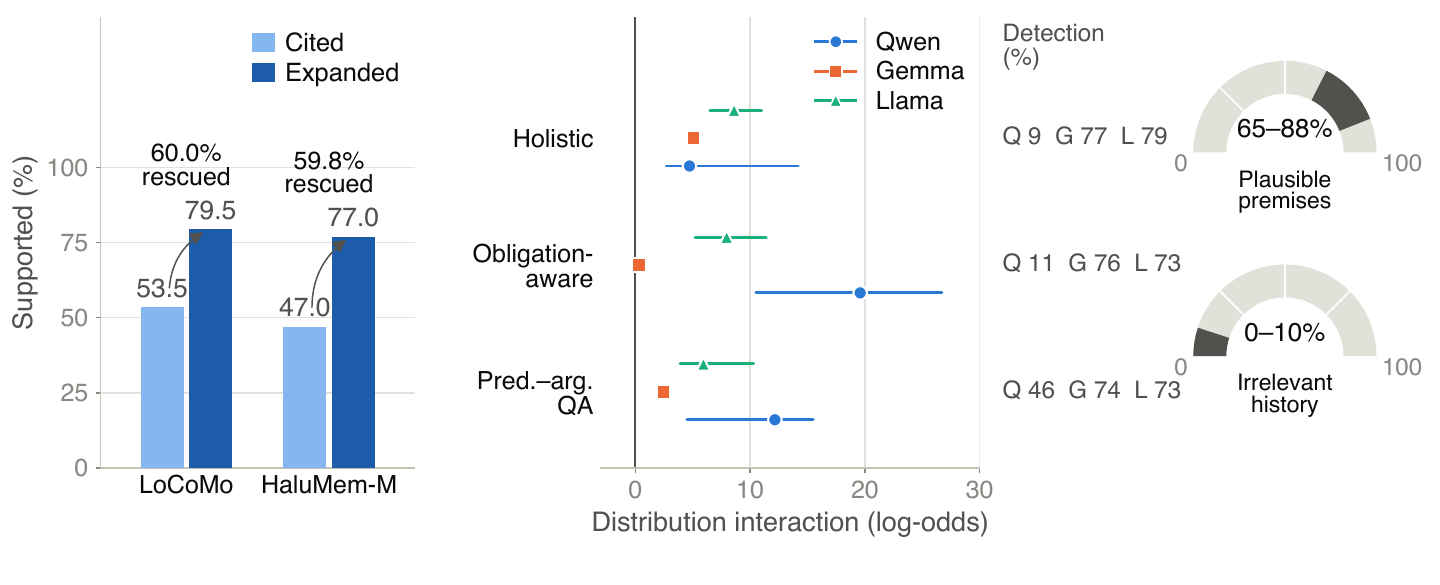}
\vspace{-30pt}
\caption{
Evidence scope and compositional verification.
\textbf{(a)} Historical evidence recovers support beyond writer-supplied citations.
\textbf{(b--c)} Distributed plausible premises reduce valid--invalid separation and increase acceptance of unsupported memory compositions relative to matched irrelevant history.
}
\label{fig:figure3}
\vspace{-5pt}
\end{figure}

%% file: figures_tex/figure5.tex
\begin{figure}[t]
\centering
\includegraphics[width=\linewidth]{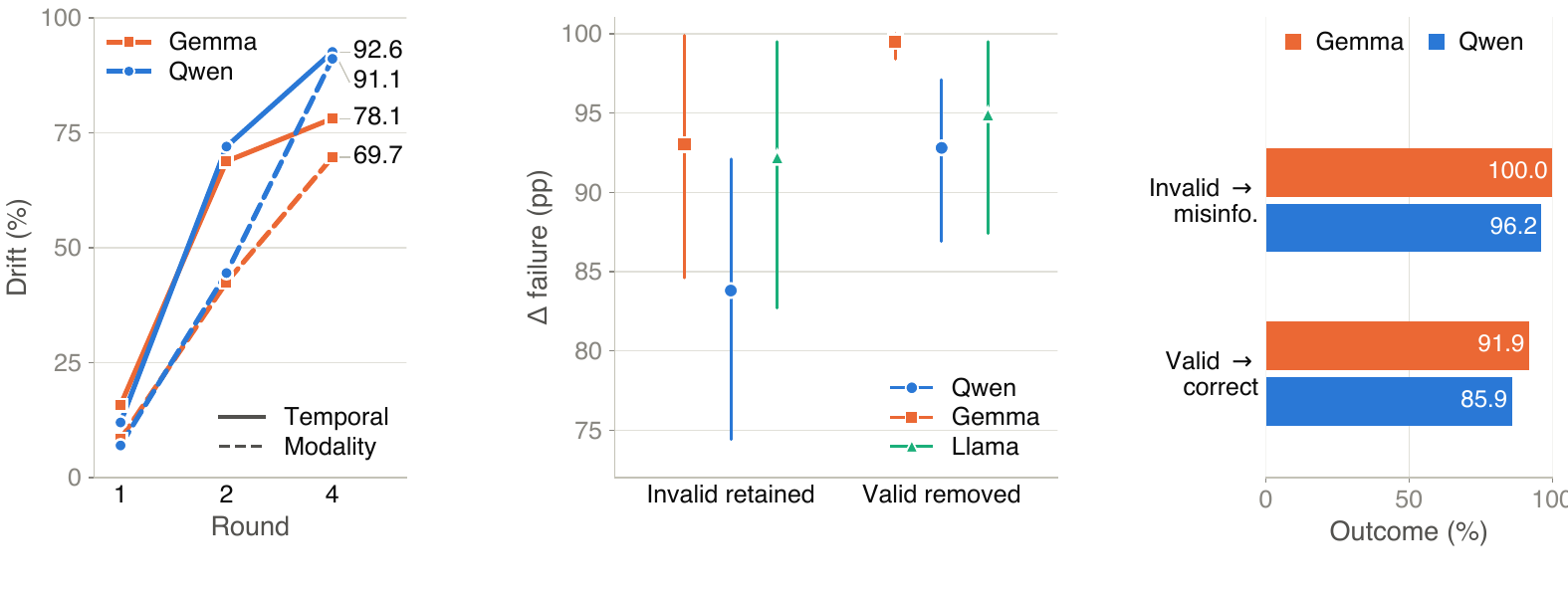}
\vspace{-30pt}
\caption{
From persistent memory to downstream behavior.
\textbf{(a)} Semantic qualifications drift during repeated consolidation.
\textbf{(b)} Distorted and missing memories both increase downstream failure.
\textbf{(c)} Natural retrieval links invalid memory to misinformation and valid memory to correct answers.
}
\label{fig:figure5}
\vspace{-5pt}
\end{figure}

%% file: figures_tex/figure6.tex
\begin{figure}[!t]
\centering
\includegraphics[width=\linewidth]{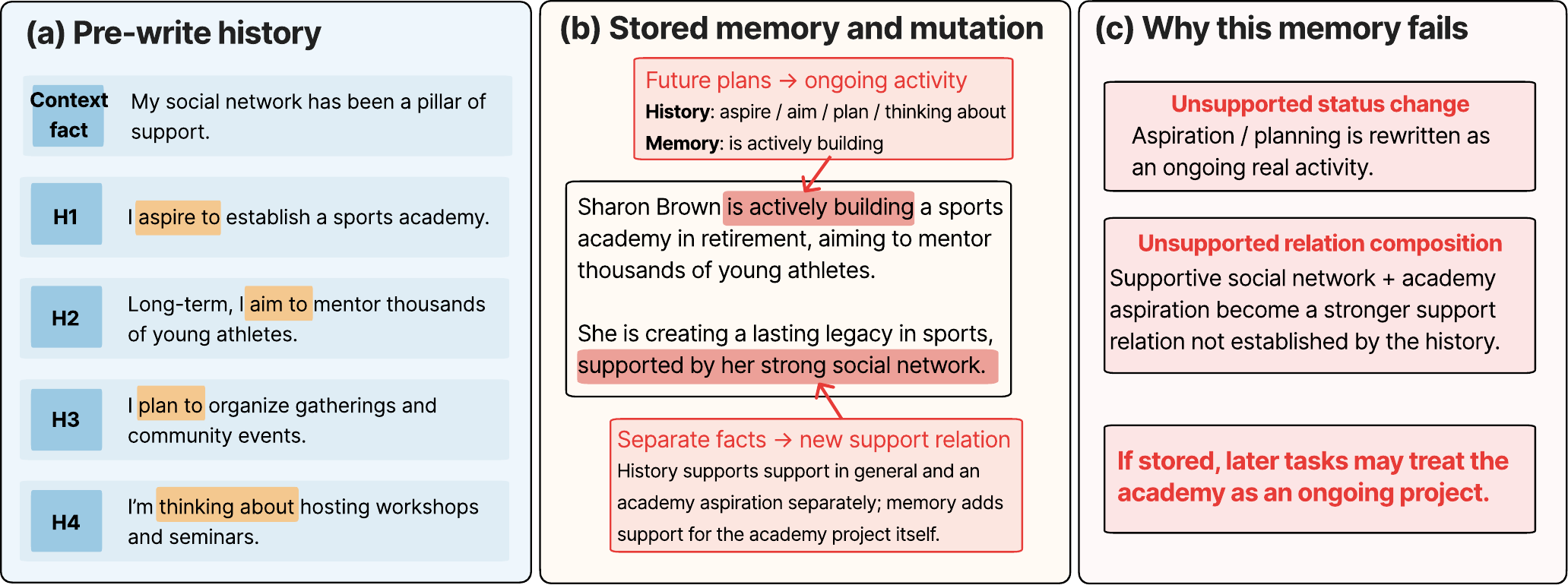}
\vspace{-20pt}
\caption{
\textbf{A natural failure of memory derivation.}
The writer compresses future-oriented plans into an ongoing activity and combines separately supported facts into a relation that the pre-write history does not establish.
}
\label{fig:natural-case}
\vspace{-15pt}
\end{figure}

%% file: paper/05_conclusion.tex
\section{Conclusion and Limitations}

We study persistent agent memory as a derivation from interaction history into persistent records that later tasks may reuse.
Across natural memory writes and controlled interventions, we identify three challenges to reliable memory writing.
First, \emph{evidence scope}: broader pre-write history recovers support for nearly 60\% of memories whose citations alone are insufficient.
Second, \emph{compositional validity}: individually plausible facts can still be combined into unsupported meaning, and distributed evidence can make such compositions harder to reject, a pattern we call the \emph{distributed-evidence paradox}.
Third, \emph{admission reliability}: recovering missing provenance substantially improves retention of provenance-repaired memories, but rejection of unsupported writes remains inconsistent across verification models, while more detailed checking can itself reject valid memories.
Downstream interventions further show that both distorted and missing memories can substantially affect later answers.
Together, these findings show that reliable memory writing requires more than finding relevant evidence: verification must also determine whether that evidence supports the full meaning that enters persistent memory.

\textbf{Limitations.} Reference labels rely primarily on two independent model assessments, with disagreements resolved by blinded adjudication, although stratified human review shows 96.0\% agreement.
Evidence expansion is retrieval-bounded, so failure to recover support does not establish historical invalidity.
Natural data contain relatively few contradicted cases and cases involving explicit relations, while our controlled suites cover only a subset of the ways meaning can be altered during memory writing.
Results also vary across verification models, verification views, prompting, and aggregation rules.
Finally, consolidation results are observational, and downstream interventions do not capture a fully closed-loop memory system.
We therefore treat these findings as evidence about the evaluated settings rather than universal claims about memory verification.

%% file: paper/appendix.tex
\clearpage
\appendix



\section{Formal Boundaries and Proofs}
\label{app:proofs}

\subsection{Evidence and information boundaries}
\label{sec:boundaries}

The first boundary concerns evidence scope. A verifier that checks each evidence unit independently and aggregates by a maximum cannot accept every valid record whose support is genuinely joint.

\begin{proposition}[Single-span evidence boundary]
\label{prop:single-span}
Suppose $m$ is licensed by $S=\{e_i,e_j\}$ under $\Gamma$, while neither $e_i$ nor $e_j$ licenses $m$ alone. Any verifier whose accept decision requires $\max_k V(m,e_k)$ to exceed a threshold rejects $m$ whenever every individual score remains below that threshold.
\end{proposition}

The second boundary concerns information, not architecture. If an interface maps two candidates to identical inputs after deleting their only differing truth-conditional relation, no downstream scorer can recover that relation.

\begin{lemma}[Representation-deletion boundary]
\label{lem:deletion}
Let $m^+$ and $m^-$ share the same mention-only representation and evidence but differ in a relation that is licensed in $m^+$ and unlicensed in $m^-$. Every verifier that factors exclusively through that shared representation assigns the pair the same score.
\end{lemma}

These statements do not rule out scalar or holistic verification. For example,
\begin{equation}
V^*(m,H_t)=I_\Gamma(m;\Evidence(H_t))
\end{equation}
is scalar-valued yet semantically complete by construction. Verification can fail because its evidence scope is insufficient or because its checks omit truth-conditional components; a single output score is not itself the cause.

\paragraph{Repair requires a separate semantics.}
Logical weakening is allowed only when the original assertion entails the replacement under the declared logic. Evidence-grounded correction is a different operation: it replaces an unsupported obligation with one separately licensed by the history. For example, \textsc{completed}$(p)$ does not generally entail \textsc{planned}$(p)$, so that edit cannot be presented as logical weakening without an additional premise.

\paragraph{Proof of Proposition~\ref{prop:single-span}.}
By assumption, every individual evidence unit yields a score below the acceptance threshold, although the joint set licenses the record. The maximum of those individual scores remains below the threshold, so the verifier rejects the record. The proposition makes no claim about a verifier that scores evidence subsets jointly. \hfill$\square$

\paragraph{Proof of Lemma~\ref{lem:deletion}.}
The verifier receives identical representation and evidence inputs for $m^+$ and $m^-$. Substituting identical inputs into its factorization yields identical scores. The result does not apply when the interface preserves the differing relation or exposes the original record. \hfill$\square$

\section{Annotation Instrument and Adjudication}
\label{app:instrument}

\paragraph{Primitives, never labels.} Reference labels are produced by routing, not by typing: annotators never write a support-status label. For each obligation, an annotator answers three primitive questions in a fixed order. \textbf{D} (\emph{determinate}): does the obligation have determinate truth conditions under $\Gamma$, answered from the record alone before the evidence spans are consulted, with a reason code required for a negative answer; every reason code names a property of the record, never of the evidence, which is what keeps D answerable before the spans are read. \textbf{E} (\emph{entailed}): is there a subset of the spans that entails the obligation under $\Gamma$; if yes, the minimal such subset is listed. \textbf{R} (\emph{refuted}): is there a subset that entails the negation of the obligation under $\Gamma$; if yes, that subset is listed. E and R have identical scope---one obligation, all spans---and differ only in direction, so neither question can collapse into the negation of the other.

\paragraph{Routing.} The four-way label is a pure function of the primitives:

\begin{center}\small
\begin{tabular}{@{}llll@{}}
\toprule
D & E & R & label \\
\midrule
no & -- & -- & ambiguous under $\Gamma$ (D's reason code) \\
yes & yes & no & supported \\
yes & no & yes & contradicted \\
yes & no & no & insufficient evidence \\
yes & yes & yes & ambiguous under $\Gamma$ ($\Gamma$-inconsistent) \\
\bottomrule
\end{tabular}
\end{center}

Routing exists for stability: directly typed multi-way labels measured $0.69$--$0.78$ agreement stability under rubric-wording revision alone, against $0.81$--$0.91$ for the primitives. Under routing, rubric wording moves the primitives, and the label moves only as far as the primitives do.

\paragraph{Refutation is time-indexed.} A span refutes an obligation only when both concern the same subject, event, and time frame under $\Gamma$; a later state change is an update, not a contradiction. This rule is part of the frozen instrument text and is applied identically in both evidence scopes.

\paragraph{Adjudication.} Every record is annotated by two independent model instances. Disagreements---and only disagreements---go to a blinded adjudicator that sees both primitive sets without identities or provenance and itself writes primitives, never labels. A record-level verdict is then routed from its obligation labels by the pre-registered precedence contradicted $>$ unsupported $>$ ambiguous $>$ supported. Two clerical defects found during review (one omitted obligation, one transposed obligation id) were repaired as transcription by full re-annotation from a fresh instance, never by editing labels.

\section{Natural-Audit Accounting and Pilot Gate}
\label{app:p08}

\paragraph{Attrition.} Every unit between the issued scaffold and the agreement coefficients reported in Section~\ref{sec:experiments} is accounted for. An obligation marked spurious by either annotator leaves the support-status family, since there is no pair left to compare.

\begin{center}\small
\begin{tabular}{@{}lr@{}}
\toprule
records annotated by both & 160 \\
scaffold obligations issued & 643 \\
\quad judgements if none were spurious ($\times 2$) & 1286 \\
\quad spurious marks & $-20$ \\
obligation judgements & 1266 \\
\midrule
obligations dropped, both annotators marked spurious & 8 \\
obligations dropped, one annotator marked spurious & 4 \\
support-status family & 631 \\
\bottomrule
\end{tabular}
\end{center}

The two drops close exactly: $8\times2+4=20$ spurious marks, and $643-8-4=631$ units. Annotator additions do not enter this family at all; they appear only in the inventory row, which is why that row's $n$ exceeds 643.

\paragraph{Four denominators for \emph{contradicted}.} The label is rare enough that the choice of denominator changes the number reported, so all four are given. Pre-adjudication: 9 of 1266 judgements (0.71\%); 6 of 631 units where either annotator used it (0.95\%); 3 of 631 where both did (0.48\%). Post-adjudication: 4 of 631 final labels (0.63\%). The reliability coefficients reported in Section~\ref{sec:experiments}
are pre-adjudication by construction; any prevalence statement uses the post-adjudication count.

\paragraph{The pilot gate.} The pre-registered gate required all four labels to occur in a 12-record pilot carrying 102 obligation judgements. At the base rate the main round subsequently measured ($p=0.0071$), the expected count of \emph{contradicted} in the pilot was $102p=0.72$ and $P(N=0)\approx e^{-0.72}=0.487$. A gate demanding the occurrence of a label below 1\% in a sample that size fails roughly half the time on a correctly functioning instrument. Nothing about the instrument was altered in response: prompts, routing thresholds and the instrument hash are identical before and after. The evidence that refutation is reachable comes from an independent 12-item probe with stipulated gold, whose three refutation items all routed to \emph{contradicted}; that probe is logically prior to the main round and does not use it. We record the gate as a protocol deviation, keep every rare-label result descriptive, and note the lesson that generalises: a pre-registered occupancy check must be powered for the rate of the rarest label it demands, or it manufactures failures that then need explaining away.

\section{Evidence Scopes and Retrieval Configuration}
\label{app:retrieval}

\paragraph{Two audits of the same records.} Every natural record is audited twice under the same instrument. The citation-only audit presents the record with its writer-supplied cited spans. The expanded-history audit presents the same record under a new, unlinkable item id with evidence equal to the cited spans united with the BM25 top-12 passages retrieved from the \emph{pre-write} history---everything up to the last span position the writer could see---capped at 16 spans total. Spans are shown in chronological order with no marking of which were cited and which retrieved; provenance and all construction flags live only in a key file the annotator never sees.

\paragraph{Unresolved is its own outcome.} Retrieval is an operational approximation to the full pre-write evidence, so its failures must not be converted into semantic verdicts. Records whose retrieval hit the search cap are flagged \texttt{search\_limited} in the key, and an expanded-scope \emph{insufficient} verdict on such a record is reported as unresolved---reference-limited, not unsupported-by-history. This rule is applied mechanically from the flag, never by judgement.

\paragraph{Prevalence estimands.} The headline citation-insufficiency rates are \emph{conversation-balanced}: each conversation contributes equally, so a single prolific conversation cannot dominate the estimate. The \emph{write-weighted} alternative, a Horvitz--Thompson estimate weighting conversations by their number of writes, moves every reported rate by at most $0.017$ and is reported alongside. Proportions carry Wilson intervals; contrasts across scopes are record-paired, with conversation-resampled intervals where the cluster structure matters.

\section{Controlled Suite Construction}
\label{app:construction}

\paragraph{Matched families.} Each of the 70 validated seed families instantiates the four cells $\{\text{local},\text{distributed}\}\times\{\text{valid},\text{invalid}\}$ over one candidate memory pair. A cell's evidence block contains its target span or spans, the family's two fixed distractors, and deterministic neutral padding drawn from the same conversation's real history up to a common evidence budget, so the only cross-cell differences are where the licensing premises sit and whether they license. Families are matched on semantic type, obligation count, candidate length, and distractors. Truth is by construction; nothing visible in the prompt names a cell or a label, construction labels live only in the key file, and item files are leak-checked mechanically with word-boundary matching.

\paragraph{Licensing-path interventions.} The remote-evidence conditions operate on support sets rather than individual spans: the label-flipping intervention removes \emph{every} sufficient support set for the target obligation, the label-preserving intervention removes only redundant evidence and retains at least one sufficient set, and the correction condition adds a temporally applicable explicit refutation. The mechanism-control variant presents the same invalid record with its plausible distributed premises replaced by irrelevant conversation content matched in token length and position (one family was dropped for residual content overlap, leaving 69).

\paragraph{Equal-budget prompt search.} Each compared interface receives the identical budget: three fixed paraphrase templates crossed with $\{0,2,4\}$ few-shot examples, nine configurations per interface. Shot examples come only from four families excluded from scoring. The winning configuration per interface is selected on development conversations by the declared metric and frozen, then scored once on untouched test conversations. Within-interface development ranges span $2$--$3.5\times$ (holistic $+21$ to $+53$; QA $+15$ to $+54$), larger than most between-interface gaps under default prompts; the faithful-atomic interface's best of nine configurations remains below the other interfaces' two-shot means, so the decomposition deficit survives its entire budget.

\section{Threshold Selection and Decision Rules}
\label{app:thresholds}

\paragraph{Verdict extraction.} Every verifier decision is read from the true token log-probabilities of the verdict words rather than from sampled text. The serving window is widened to the top 100 log-probabilities, and the request is rejected rather than silently truncated if the server cannot honour it. Two quantities are recorded for each judgement: the normalised probability $p(\textit{yes})/(p(\textit{yes})+p(\textit{no}))$, which saturates against $1.0$, and the log-odds $\log p(\textit{yes})-\log p(\textit{no})$, which ranks identically but keeps near-ties apart. All thresholding and paired analyses operate on log-odds. A record for which neither verdict token appears in the window is excluded and counted as a scoring error, never defaulted to mid-scale; a record for which only one token appears yields a real score with an unmeasurable margin and is counted separately as a rail hit.

\paragraph{The rail audit.} On Gemma, 84--99\% of scores sit on the window floor. A full-vocabulary echo audit on 120 Gemma rows (stratified over rail direction) and 33 Qwen rows recovered the losing verdict's exact probability: median $10^{-10}$, maximum $4.2\times10^{-9}$. Replacing the floored values with exact probabilities flips 0 of 120 Gemma decisions and 1 of 33 Qwen decisions (a $+0.031$-margin record inside the known rerun variation). Near-binary verdicts are therefore a property of the model, not of the scoring pipeline.

\paragraph{Backbone-specific decision rules.} The graded backbone (Qwen) is thresholded on log-odds with leave-one-conversation-out cross-fitting: thresholds are selected on the held-in conversations at a selection false-positive rate of at most $0.10$ and every reported number is evaluated out-of-fold. The dense backbones produce near-binary score distributions that cannot support a tuned threshold, so Gemma and Llama are read at their own argmax operating point, $p(\textit{yes})>p(\textit{no})$, with nothing fitted. The same asymmetry appears in deployment: at the untuned argmax point the graded backbone gates almost nothing on natural records (96--99\% false admission, with only 2 of 400 records inside $|\text{log-odds}|<1$), while the dense backbones' argmax points already gate. Comparisons are therefore made within a backbone, never across operating points.

\paragraph{Paired uncertainty.} Policy contrasts on the natural records are paired at the record level. We report exact McNemar tests on the discordant pairs, together with $10{,}000$-replicate record-paired and conversation-clustered bootstrap intervals; the clustered intervals are the ones quoted in the main text. For the admission contrast the discordant split is $12{:}6$ on 78 negative records, which is why that reduction is reported as directionally consistent but unresolved.

\paragraph{Determinism.} Every scoring job runs two identical passes and records the gap rather than asserting bit-equality: median absolute log-odds difference per rerun is $\approx 0.2$--$0.5$ on the mixture-of-experts backbone and $\approx 10^{-8}$ on the dense backbones. The one Qwen rail-audit flip above sits inside this recorded variation.

\section{Robustness and Protocol Controls}
\label{app:controls}

This appendix collects the controls referenced in Section~\ref{sec:analysis}. Table~\ref{tab:representation-robustness} varies how the same candidate memory and evidence are rendered: correct, isomorphic, and shuffled composition graphs, length-matched filler, and compact relations. Structure helps or hurts depending on backbone and rendering, and a shuffled graph can dominate the judgement entirely, so no representation is a universal repair. Table~\ref{tab:protocol-controls} reports the protocol-level controls: oracle upstream inputs, the equal-budget prompt search under which the top three interfaces converge to within four points while the winning interface flips across backbones, alternative aggregation rules over separately scored obligations (minimum $0.266$, mean $0.101$, noisy-AND $0.304$ unsupported detection on Qwen---the operating point moves, the tension remains), and verification compute.

Table~\ref{tab:filler-noise} gives the full accumulated-verification-noise results behind Section~\ref{sec:analysis}: adding obligations independently verified as true lowers support scores monotonically on every backbone, with backbone-scaled magnitude. Table~\ref{tab:remote-controls} shows the two controls that locate the distributed-evidence paradox's mechanism: the verifier notices in-input support deletion and honours applicable corrections, and the paradox disappears when plausible premises are replaced by token-matched irrelevant content---the failure is plausible-premise composition bias, not context length or neglect of distant evidence.

\input{Tables/represent_robust}

\input{Tables/more_robust}

\input{Tables/filler_noise}

\input{Tables/remote_controls}

\section{Downstream Causal Protocol}
\label{app:cace}

\paragraph{Design.} The downstream probe changes exactly one memory slot available to a reader---correct, distorted, or absent---while holding the slot's retrieval position and all surrounding context fixed. Each example is evaluated on two queries: the \emph{harm} query, whose answer the distortion changes, and the \emph{content} query, which the record uniquely answers. Two causal quantities are computed, paired within example and never averaged: $\mathrm{CACE}_{\text{harm}} = P(\text{fail}\mid do(\text{distorted})) - P(\text{fail}\mid do(\text{correct}))$ on the harm query (the cost of false admission) and $\mathrm{CACE}_{\text{reject}} = P(\text{fail}\mid do(\text{absent})) - P(\text{fail}\mid do(\text{correct}))$ on the content query (the cost of false rejection). All six answer-option permutations are run and averaged so option position cannot carry the effect, with three replicates at temperature zero capturing server nondeterminism only.

\paragraph{Control floors gate every causal number.} No causal quantity is read unless four control floors pass at $0.80$, all-or-nothing: the reader must answer the content query correctly with the correct record present, must say the record is absent when it is absent, must recall correctly, and must not manufacture reasons. These floors can genuinely fail---an always-name policy fails the no-recall floor and an always-reason policy fails the no-reason floor---and every backbone passed all gating floors. Malformed responses are excluded as errors, never counted as data. The off-diagonal cells (distortion evaluated on the content query, removal on the harm query) are reported as specificity checks, not pooled into either estimate: a swapped holder makes the gold statement unrecorded and is expected to move on the content query, while a spurious connective leaves it recorded and is not.

\paragraph{Results across backbones.} $\mathrm{CACE}_{\text{harm}}$ is $+0.838$ (Qwen), $+0.930$ (Gemma), $+0.922$ (Llama); $\mathrm{CACE}_{\text{reject}}$ is $+0.928$, $+0.995$, $+0.949$. Persistent two-sided consequences are a property of the setting, not of a model class. One profile difference is worth recording: with a distorted record on the content query, Llama abstains (``records neither'') at $0.505$ rather than answering.

\section{Additional Natural Cases}
\label{app:cases}

Table~\ref{tab:natural-cases} complements the aspiration-to-activity case in Section~\ref{sec:case-study} with four further unedited writes from the truth-labelled natural set, covering event-status mutation, cross-entity re-binding of durations, attributes, and causes, and a time-indexed refutation.

\input{Tables/natural_cases}

\section{Verification Interface Prompts}
\label{app:prompts}

All interfaces share one scoring harness, one evidence rendering, and one verdict extraction; only the prompt below changes. Multi-unit interfaces score each unit with the per-unit template and aggregate by the minimum, since a record is licensed only if every obligation is. Placeholders in braces are filled per record.

\paragraph{Holistic.}
\begin{quote}\small\ttfamily
You are checking whether a memory record is licensed by conversation evidence.\\[2pt]
RECORD: \{record\}\\
EVIDENCE (\{n\} spans): \{evidence\}\\[2pt]
Is the record as a whole licensed by this evidence? A record is licensed only if every claim it makes, and every relation it asserts between claims, is supported. If it asserts a relation, the evidence must license the relation itself, not merely each side separately.\\
Answer with one word, yes or no.
\end{quote}

\paragraph{Atomic claims (per unit).}
\begin{quote}\small\ttfamily
You are checking whether ONE claim from a memory record is licensed by conversation evidence.\\[2pt]
CLAIM: \{unit\}\\
EVIDENCE (\{n\} spans): \{evidence\}\\[2pt]
Is this claim licensed by the evidence? Judge only this claim, not the rest of the record.\\
Answer with one word, yes or no.
\end{quote}

\paragraph{Residual (composition beyond listed claims).}
\begin{quote}\small\ttfamily
You are checking a memory record for anything it asserts BEYOND the individual claims listed below.\\[2pt]
RECORD: \{record\}\\
CLAIMS ALREADY CHECKED SEPARATELY: \{parts\}\\
EVIDENCE (\{n\} spans): \{evidence\}\\[2pt]
Assume each listed claim is licensed. The question is only about what the record asserts in addition: a relation it draws between claims, a modifier or circumstance it attaches, or a binding of an attribute to a particular person. Is everything the record asserts beyond the listed claims licensed by the evidence?\\
Answer with one word, yes or no.
\end{quote}

\paragraph{Composition graph.}
\begin{quote}\small\ttfamily
You are checking whether a memory record is licensed by conversation evidence.\\[2pt]
RECORD: \{record\}\\
ITS STRUCTURE: claims it makes: \{claims\} plus whatever the record asserts to connect, modify or bind them\\
EVIDENCE (\{n\} spans): \{evidence\}\\[2pt]
The record is licensed only if every claim above is licensed AND everything the record asserts to connect, modify or bind those claims is licensed. Evidence that supports the claims separately does not by itself license a relation drawn between them, a circumstance attached to them, or which person an attribute belongs to.\\
Answer with one word, yes or no.
\end{quote}

\paragraph{Compact relations.}
\begin{quote}\small\ttfamily
You are checking whether a memory record is licensed by conversation evidence.\\[2pt]
The record, written out as the claims it makes and the links it draws between them: \{relations\}\\
EVIDENCE (\{n\} spans): \{evidence\}\\[2pt]
Every line above must be licensed for the record to be licensed, including the lines that link claims together. Evidence supporting the linked claims separately does not license a link drawn between them.\\
Answer with one word, yes or no.
\end{quote}

\paragraph{Obligation-aware joint (admission gate).}
\begin{quote}\small\ttfamily
You are checking whether a memory record is licensed by conversation evidence.\\[2pt]
RECORD: \{record\}\\
The record asserts, at minimum, each of these obligations: \{obligations\}\\
EVIDENCE (\{n\} spans): \{evidence\}\\[2pt]
Considering every obligation against the evidence together: is the record licensed by the evidence? Answer with one word, yes or no.
\end{quote}

\paragraph{Predicate--argument QA.} The QA interface derives predicate--argument questions from the record and scores each question with the full evidence block; both decision flows are evaluated: one joint verdict over all questions, and the faithful per-question variant with minimum aggregation. Every call in every interface sees the identical evidence block---decomposing interfaces restrict the claim, never the evidence.

\section{Implementation and Reporting}
\label{app:reporting}

\paragraph{Experiment settings.} Table~\ref{tab:evaluation-suites} records the writer, verifier, and reader configurations, serving parameters, and evidence budgets used by each experiment.

\input{Tables/experiments_settings_table}

\paragraph{Reporting checklist.} Every result table must report sample size, sampling frame, denominator, history-level split, threshold-selection split, actual test FPR, model and prompt version, retrieval scope and budget, number of development trials, cluster unit, interval construction, tie rate, latency, calls, and tokens where applicable. Within-pair ranking and standard ROC-AUC are named and reported separately. Unclipped label log-probability differences are used to audit score saturation; the number of unique scores and pairwise ties accompanies each ranking result.

%% file: Tables/represent_robust.tex
\begin{table*}[t]
\centering
\caption{
Representation robustness on fixed candidate memories and evidence.
Values are unsupported-relation detection / multi-span valid retention.
}
\label{tab:representation-robustness}

\scriptsize
\setlength{\tabcolsep}{3.2pt}
\renewcommand{\arraystretch}{0.90}

\begin{tabular*}{\textwidth}{
@{\extracolsep{\fill}}
llccc
@{}
}
\toprule
\textbf{Model}
&
\textbf{Representation}
&
\shortstack{\textbf{Unsupported}\\\textbf{Detection}$\uparrow$}
&
\shortstack{\textbf{Multi-span}\\\textbf{Retention}$\uparrow$}
&
\textbf{Observation}
\\
\midrule

\multirow{6}{*}{\textbf{Qwen}}
& Raw memory
& 0.608 & 0.846
& holistic baseline \\

& Correct composition graph
& 0.468 & 0.923
& no stable gain \\

& Isomorphic graph
& 0.354 & 0.974
& rendering-sensitive \\

& Shuffled graph
& 0.430 & 0.846
& graph form alone does not help \\

& Length-matched filler
& 0.544 & 0.846
& length alone does not explain effect \\

& Compact relations
& 0.595 & 0.949
& strong retention / detection balance \\

\midrule

\multirow{6}{*}{\textbf{Gemma}$^\dagger$}
& Raw memory
& 0.937 & 0.923
& holistic baseline \\

& Correct composition graph
& 0.924 & 0.333
& retention drops \\

& Isomorphic graph
& 0.924 & 0.154
& rendering-sensitive \\

& Shuffled graph
& \multicolumn{2}{c}{505 / 512 records rejected}
& structure can dominate judgment \\

& Length-matched filler
& 0.734 & 0.923
& length alone does not explain effect \\

& Compact relations
& \multicolumn{2}{c}{all 39 multi-span memories rejected}
& degenerate operating point \\

\midrule

\multicolumn{5}{@{}l}{\textit{Llama baseline-only transfer}} \\

\textbf{Llama}$^\dagger$
& Holistic
& 0.595 & 1.000
& valid rejection = 0.016 \\

\textbf{Llama}$^\dagger$
& Atomic claims
& 0.000 & 1.000
& valid rejection = 0.000 \\

\bottomrule
\end{tabular*}

\vspace{1mm}
\begin{minipage}{\textwidth}
\scriptsize
$^\dagger$Gemma and Llama use their argmax operating points.
Llama derivation-aware representation variants were not run in this suite.
\end{minipage}

\vspace{-3mm}
\end{table*}

%% file: Tables/more_robust.tex
\begin{table*}[t]
\centering
\caption{
Protocol controls.
Correct upstream inputs, prompt search, alternative aggregation, and additional
verification compute change the operating point but do not provide a universal solution.
}
\label{tab:protocol-controls}

\scriptsize
\setlength{\tabcolsep}{3.0pt}
\renewcommand{\arraystretch}{0.90}

\begin{tabular*}{\textwidth}{
@{\extracolsep{\fill}}
lccccc
@{}
}
\toprule

\multicolumn{6}{@{}l}{\textit{Oracle upstream inputs}} \\
\midrule

\textbf{Setting}
& \textbf{Model}
& \textbf{Full}
& \textbf{Atom}
& \textbf{Graph}
& \textbf{Metric}
\\
\midrule

Gold decomposition + gold evidence
& Qwen
& 0.292
& 0.555
& 0.394
& unsupported detection
\\

\midrule

\multicolumn{6}{@{}l}{\textit{Equal-budget prompt search}} \\
\midrule

\textbf{Interface}
& \multicolumn{2}{c}{\textbf{Qwen}}
& \multicolumn{2}{c}{\textbf{Llama}}
& \textbf{Observation}
\\
\cmidrule(lr){2-3}
\cmidrule(lr){4-5}

&
\textbf{Default}
& \textbf{Tuned}
& \textbf{Default}
& \textbf{Tuned}
& \\

\midrule

Holistic
& 12.8 & 46.2
& 55.4 & 61.7
& interface gap narrows
\\

Obligation-aware
& 28.8 & 47.2
& 57.5 & 66.7
& ranking changes
\\

Predicate--argument QA
& 37.6 & 48.1
& 66.6 & 58.8
& ranking changes
\\

\midrule

\multicolumn{6}{@{}l}{\textit{Aggregation over separately scored obligations}} \\
\midrule

\textbf{Model}
& \textbf{Minimum}
& \textbf{Mean}
& \textbf{Noisy-AND}
& \multicolumn{2}{c}{\textbf{Metric}}
\\
\midrule

Qwen
& 0.266
& 0.101
& \best{0.304}
& \multicolumn{2}{c}{unsupported detection}
\\

\midrule

\multicolumn{6}{@{}l}{\textit{Verification compute}} \\
\midrule

\textbf{Interface}
& \textbf{Model}
& \textbf{Calls}
& \textbf{Input tokens}
& \textbf{Detection}
& \textbf{Relative cost}
\\
\midrule

Holistic
& Qwen
& 512
& 201,766
& 0.608
& $1.0\times$
\\

Per-obligation
& Qwen
& 1,106
& 436,340
& 0.291
& $2.2\times$
\\

\bottomrule
\end{tabular*}

\vspace{1mm}
\begin{minipage}{\textwidth}
\scriptsize
\textit{Notes.}
Oracle-input results use a separate 235-member minimal-pair suite and should not
be numerically compared with the 512-record main verification matrix.
Equal-budget prompt search uses the same nine-configuration search budget for
each interface.
\end{minipage}

\vspace{-3mm}
\end{table*}

%% file: Tables/filler_noise.tex
\begin{table*}[t]
\centering
\caption{
Accumulated verification noise.
Supported memories receive $k$ additional obligations independently verified
as true; the candidate memory is unchanged, so any score decrease measures
noise from checking more correct content, not detection of a defect.
}
\label{tab:filler-noise}

\scriptsize
\setlength{\tabcolsep}{3.0pt}
\renewcommand{\arraystretch}{0.90}

\begin{tabular*}{\textwidth}{
@{\extracolsep{\fill}}
lcccccc
@{}
}
\toprule
\textbf{Model}
&
\shortstack{\textbf{Median} $\Delta$\textbf{log-odds}\\$k{=}2$}
&
\shortstack{\textbf{Median} $\Delta$\textbf{log-odds}\\$k{=}4$}
&
\shortstack{\textbf{Records moving}\\\textbf{down} ($k{=}4$)}
&
\shortstack{\textbf{Acceptance}\\$k{=}0$}
&
\shortstack{\textbf{Acceptance}\\$k{=}2$}
&
\shortstack{\textbf{Acceptance}\\$k{=}4$}
\\
\midrule

\textbf{Qwen}
& $-5.7$ & $-7.7$ & 98.7\%
& 0.994 & 0.987 & 0.981 \\

\textbf{Gemma}$^\dagger$
& $-43.3$ & $-45.9$ & 100.0\%
& 0.933 & 0.121 & 0.026 \\

\textbf{Llama}$^\dagger$
& -- & $-20.0$ & 99.0\%
& 0.974 & -- & 0.355 \\

\bottomrule
\end{tabular*}

\vspace{1mm}
\begin{minipage}{\textwidth}
\scriptsize
\textit{Notes.}
Fillers are drawn from other records of the same conversation whose
expanded-scope verdict is supported and whose every resolved obligation was
entailed; the $k{=}2$ list is a prefix of the $k{=}4$ list, so $k$ is the only
difference between conditions.
The direction is universal and the magnitude is backbone-scaled;
on the graded backbone the binary acceptance moves little while the underlying
log-odds fall, which is why threshold-based deployments of the same verifier
can behave very differently.
$^\dagger$Argmax operating point. ``--'': condition not run on this backbone.
\end{minipage}

\vspace{-3mm}
\end{table*}

%% file: Tables/remote_controls.tex
\begin{table*}[t]
\centering
\caption{
Remote-history and mechanism controls (family-paired, Qwen verifier).
Top: when the manipulation is inside the verifier's input, the verifier is
competent---deleted support is noticed, redundant-path deletion never flips a
decision, and a temporally applicable correction is honoured.
Bottom: the same invalid record is accepted when its plausible premises are
present but rejected when they are replaced by token-matched irrelevant
history, isolating plausible-premise composition bias as the mechanism of the
distributed-evidence paradox.
}
\label{tab:remote-controls}

\scriptsize
\setlength{\tabcolsep}{3.0pt}
\renewcommand{\arraystretch}{0.90}

\begin{tabular*}{\textwidth}{
@{\extracolsep{\fill}}
lcccc
@{}
}
\toprule

\multicolumn{5}{@{}l}{\textit{Remote-evidence interventions (69 families, five conditions)}} \\
\midrule

\textbf{Interface}
&
\shortstack{\textbf{Accept}\\\textbf{intact}$\uparrow$}
&
\shortstack{\textbf{Still accept after}\\\textbf{support deletion}$\downarrow$}
&
\shortstack{\textbf{Wrong flip after}\\\textbf{redundant deletion}$\downarrow$}
&
\shortstack{\textbf{False admit after}\\\textbf{applicable correction}$\downarrow$}
\\
\midrule

Holistic
& 0.986 & 0.116 & 0.000 & 0.014 \\

Obligation-aware
& 1.000 & 0.000 & 0.000 & 0.000 \\

Predicate--argument QA
& 1.000 & 0.000 & 0.000 & 0.000 \\

\midrule

\multicolumn{5}{@{}l}{\textit{Mechanism control: plausible premises versus token-matched irrelevant content}} \\
\midrule

\textbf{Interface}
&
\shortstack{\textbf{Accept with}\\\textbf{plausible premises}}
&
\shortstack{\textbf{Accept with}\\\textbf{irrelevant content}}
&
\multicolumn{2}{c}{\shortstack{\textbf{Paired} $\Delta$\textbf{log-odds} [95\% CI]}}
\\
\midrule

Holistic
& 0.884 & 0.101
& \multicolumn{2}{c}{$+21.9$ [$+19.0,+25.5$]} \\

Obligation-aware
& 0.884 & 0.000
& \multicolumn{2}{c}{$+29.9$ [$+27.7,+31.4$]} \\

Predicate--argument QA
& 0.652 & 0.000
& \multicolumn{2}{c}{$+27.0$ [$+24.6,+30.7$]} \\

\bottomrule
\end{tabular*}

\vspace{1mm}
\begin{minipage}{\textwidth}
\scriptsize
\textit{Notes.}
Median paired $\Delta$log-odds: full support deletion $28$--$37$; redundant-path
deletion $0.1$--$1.4$; explicit correction $32$--$37$.
On Gemma the applicable-correction false-admit rate is $0.043$--$0.145$.
In the mechanism control both conditions present the same invalid record with
distributed spans matched in length and position; only whether the spans carry
premise content differs.
\end{minipage}

\vspace{-3mm}
\end{table*}

%% file: Tables/natural_cases.tex
\begin{table*}[t]
\centering
\caption{
Additional unedited natural writes exhibiting derivation failures.
Each row quotes the pre-write history and the resulting memory; verdicts are
the adjudicated labels under the citation-only and expanded-history audits.
The second case is the study's canonical illustration that citation matching
and history-level validity are different targets: its cited spans pass the
provenance check while the expanded audit finds the meaning unsupported.
}
\label{tab:natural-cases}

\scriptsize
\setlength{\tabcolsep}{3.0pt}
\renewcommand{\arraystretch}{1.0}

\begin{tabularx}{\textwidth}{
@{}
>{\raggedright\arraybackslash}X
>{\raggedright\arraybackslash}X
l
c
c
@{}
}
\toprule
\textbf{Pre-write history says}
&
\textbf{Memory says}
&
\textbf{Mutation}
&
\shortstack{\textbf{Cited}\\\textbf{verdict}}
&
\shortstack{\textbf{Expanded}\\\textbf{verdict}}
\\
\midrule

``I've been working on this car, doing engine swaps and suspension
modifications. Now I'm \emph{learning about} body modifications.''
&
``\ldots transforming it with engine swaps, suspension modifications,
\emph{and body modifications}.''
&
prospective $\rightarrow$ performed
&
unsup.
&
unsup.
\\
\addlinespace

``What instrument are you playing?'' --- ``I'm learning how to play the
\emph{violin} now\ldots'' / ``How long have you been playing the
\emph{piano} again?'' --- ``I've been playing for about \emph{four months}.''
&
``Tim has been learning to play the \emph{violin} for about \emph{four
months}\ldots''
&
duration re-bound across entities
&
supp.
&
unsup.
\\
\addlinespace

``I'm still just learning how to draw, but I love expressing myself
\emph{through writing}.'' / ``I've been a bit \emph{frustrated} lately with
my \emph{new phone}.''
&
``Sam has recently taken up drawing as a new form of \emph{self-expression}
\ldots despite occasional \emph{frustration with his progress}.''
&
attribute and cause re-bound
&
unsup.
&
unsup.
\\
\addlinespace

``I scored a deal to \emph{continue} collaboration with Frank Ocean!''
&
``Calvin expressed excitement about his \emph{new} collaboration with Frank
Ocean\ldots''
&
ongoing $\rightarrow$ new (refuted)
&
contra.
&
contra.
\\

\bottomrule
\end{tabularx}

\vspace{1mm}
\begin{minipage}{\textwidth}
\scriptsize
\textit{Notes.}
All four records are unedited writer outputs from the truth-labelled natural
set; none was constructed for this table.
Emphasis added.
Together with the aspiration-to-activity case in the main text, these span the
mutation families the obligation decomposition targets: event status
(prospective versus performed), binding (attributes, durations, and causes
attached to the wrong entity), and time-indexed refutation.
\end{minipage}

\vspace{-3mm}
\end{table*}

%% file: Tables/experiments_settings_table.tex
\begin{table*}[t]
\centering
\caption{
Evaluation suites used to study evidence scope, compositional validity,
admission reliability, and later memory use.
}
\label{tab:evaluation-suites}

\scriptsize
\setlength{\tabcolsep}{3.0pt}
\renewcommand{\arraystretch}{0.90}

\begin{tabularx}{\textwidth}{
@{}
l
>{\raggedright\arraybackslash}X
c
>{\raggedright\arraybackslash}X
@{}
}
\toprule
\textbf{Evaluation}
&
\textbf{Agent-memory question}
&
\textbf{Scale}
&
\textbf{Primary comparison}
\\
\midrule

Natural paired audit
&
Is a memory unsupported, or is writer provenance incomplete?
&
400 writes
&
Cited evidence $\rightarrow$ expanded pre-write evidence
\\

Natural admission replay
&
Which supported and unsupported memories survive write-time verification?
&
391 writes
&
Citation-only / expanded history / composition-aware
\\

Matched distribution
&
Can a verifier distinguish licensed memory from plausible composition when
support is distributed?
&
70 families $\times$ 4
&
Local/distributed $\times$ valid/invalid
\\

Semantic verification
&
Which representation best preserves relations asserted by a memory?
&
512 records
&
Prior paradigms vs.\ derivation-aware interfaces
\\

Remote-history controls
&
Are failures caused by ignored history or by plausible but non-licensing premises?
&
Matched interventions
&
Necessary-path removal / redundant-path removal / correction
\\

Oracle diagnostics
&
Do errors remain with correct evidence and decomposition?
&
235 records
&
Gold evidence + gold decomposition
\\

Supported-filler intervention
&
Can additional correct checks cause valid memory to be rejected?
&
391 labelled records
&
$k=0,2,4$ verified-true added obligations
\\

Natural gate replay
&
Does verification exert selection pressure on naturally written memory?
&
663 writes
&
Single-span licensor present / absent
\\

Memory lifecycle probes
&
How do distorted or missing memories affect later agent behavior?
&
39 pairs + 35 retrieval items
&
Correct / distorted / absent memory
\\

\bottomrule
\end{tabularx}

\vspace{-3mm}
\end{table*}

%% file: iclr2027_conference.bib
@misc{packer2023memgpt,
      title={MemGPT: Towards LLMs as Operating Systems}, 
      author={Charles Packer and Sarah Wooders and Kevin Lin and Vivian Fang and Shishir G. Patil and Ion Stoica and Joseph E. Gonzalez},
      year={2024},
      eprint={2310.08560},
      archivePrefix={arXiv},
      primaryClass={cs.AI},
      url={https://arxiv.org/abs/2310.08560}, 
}

@misc{chhikara2025mem0,
      title={Mem0: Building Production-Ready AI Agents with Scalable Long-Term Memory}, 
      author={Prateek Chhikara and Dev Khant and Saket Aryan and Taranjeet Singh and Deshraj Yadav},
      year={2025},
      eprint={2504.19413},
      archivePrefix={arXiv},
      primaryClass={cs.CL},
      url={https://arxiv.org/abs/2504.19413}, 
}

@misc{wu2025longmemeval,
      title={LongMemEval: Benchmarking Chat Assistants on Long-Term Interactive Memory}, 
      author={Di Wu and Hongwei Wang and Wenhao Yu and Yuwei Zhang and Kai-Wei Chang and Dong Yu},
      year={2025},
      eprint={2410.10813},
      archivePrefix={arXiv},
      primaryClass={cs.CL},
      url={https://arxiv.org/abs/2410.10813}, 
}

@misc{joshi2026eywa,
      title={Eywa: Provenance-Grounded Long-Term Memory for AI Agents}, 
      author={Resham Joshi},
      year={2026},
      eprint={2605.30771},
      archivePrefix={arXiv},
      primaryClass={cs.CL},
      url={https://arxiv.org/abs/2605.30771}, 
}

@misc{zhang2026consistencygate,
      title={ConsistencyGate: Preventing Memory Contamination in LLM Agents via Self-Consistency Admission Control}, 
      author={Yan Zhang and Shibo Li},
      year={2026},
      eprint={2607.22962},
      archivePrefix={arXiv},
      primaryClass={cs.AI},
      url={https://arxiv.org/abs/2607.22962}, 
}

@misc{cui2026memtxn,
      title={MemTxn: A Transaction Boundary for Source-Supported Updates and Complete-State Recovery in Agent Memory}, 
      author={Hanshuai Cui and Zhiqing Tang and Zhi Yao and Fanshuai Meng and Qianli Ma and Weijia Jia},
      year={2026},
      eprint={2607.27834},
      archivePrefix={arXiv},
      primaryClass={cs.AI},
      url={https://arxiv.org/abs/2607.27834}, 
}

@inproceedings{latimer2026hindsight,
    title = "Hindsight: Structured Agent Memory that Retains, Recalls, and Reflects",
    author = "Latimer, Christopher  and
      Boschi, Nicol{\`o}  and
      Neeser, Andrew  and
      Bartholomew, Chris  and
      Srivastava, Gaurav  and
      Wang, Xuan  and
      Ramakrishnan, Naren",
    editor = "Durrett, Greg  and
      Jian, Ping",
    booktitle = "Proceedings of the 64th Annual Meeting of the {A}ssociation for {C}omputational {L}inguistics (Volume 3: System Demonstrations)",
    month = jul,
    year = "2026",
    address = "San Diego, California, United States",
    publisher = "Association for Computational Linguistics",
    url = "https://aclanthology.org/2026.acl-demo.27/",
    doi = "10.18653/v1/2026.acl-demo.27",
    pages = "275--285",
    ISBN = "979-8-89176-392-0"
}

@misc{jin2026memir,
      title={Mitigating Provenance-Role Collapse in Long-Term Agents via Typed Memory Representation}, 
      author={Zhengda Jin and Bingbing Wang and Jing Li and Ruifeng Xu and Min Zhang},
      year={2026},
      eprint={2605.25869},
      archivePrefix={arXiv},
      primaryClass={cs.CL},
      url={https://arxiv.org/abs/2605.25869}, 
}

@misc{maynez2020faithfulness,
      title={On Faithfulness and Factuality in Abstractive Summarization}, 
      author={Joshua Maynez and Shashi Narayan and Bernd Bohnet and Ryan McDonald},
      year={2020},
      eprint={2005.00661},
      archivePrefix={arXiv},
      primaryClass={cs.CL},
      url={https://arxiv.org/abs/2005.00661}, 
}

@misc{pagnoni2021frank,
      title={Understanding Factuality in Abstractive Summarization with FRANK: A Benchmark for Factuality Metrics}, 
      author={Artidoro Pagnoni and Vidhisha Balachandran and Yulia Tsvetkov},
      year={2021},
      eprint={2104.13346},
      archivePrefix={arXiv},
      primaryClass={cs.CL},
      url={https://arxiv.org/abs/2104.13346}, 
}

@misc{goyal2020dae,
      title={Evaluating Factuality in Generation with Dependency-level Entailment}, 
      author={Tanya Goyal and Greg Durrett},
      year={2020},
      eprint={2010.05478},
      archivePrefix={arXiv},
      primaryClass={cs.CL},
      url={https://arxiv.org/abs/2010.05478}, 
}

@misc{ribeiro2022factgraph,
      title={FactGraph: Evaluating Factuality in Summarization with Semantic Graph Representations}, 
      author={Leonardo F. R. Ribeiro and Mengwen Liu and Iryna Gurevych and Markus Dreyer and Mohit Bansal},
      year={2022},
      eprint={2204.06508},
      archivePrefix={arXiv},
      primaryClass={cs.CL},
      url={https://arxiv.org/abs/2204.06508}, 
}

@misc{cattan2026qasem,
      title={Localizing Factual Inconsistencies in Attributable Text Generation}, 
      author={Arie Cattan and Paul Roit and Shiyue Zhang and David Wan and Roee Aharoni and Idan Szpektor and Mohit Bansal and Ido Dagan},
      year={2025},
      eprint={2410.07473},
      archivePrefix={arXiv},
      primaryClass={cs.CL},
      url={https://arxiv.org/abs/2410.07473}, 
}

@misc{lewis2025vista,
      title={VISTA: Verification In Sequential Turn-based Assessment}, 
      author={Ashley Lewis and Andrew Perrault and Eric Fosler-Lussier and Michael White},
      year={2026},
      eprint={2510.27052},
      archivePrefix={arXiv},
      primaryClass={cs.CL},
      url={https://arxiv.org/abs/2510.27052}, 
}

@misc{hu2025memoryagentbench,
      title={Evaluating Memory in LLM Agents via Incremental Multi-Turn Interactions}, 
      author={Yuanzhe Hu and Yu Wang and Julian McAuley},
      year={2026},
      eprint={2507.05257},
      archivePrefix={arXiv},
      primaryClass={cs.CL},
      url={https://arxiv.org/abs/2507.05257}, 
}

@misc{min2023factscore,
      title={FActScore: Fine-grained Atomic Evaluation of Factual Precision in Long Form Text Generation}, 
      author={Sewon Min and Kalpesh Krishna and Xinxi Lyu and Mike Lewis and Wen-tau Yih and Pang Wei Koh and Mohit Iyyer and Luke Zettlemoyer and Hannaneh Hajishirzi},
      year={2023},
      eprint={2305.14251},
      archivePrefix={arXiv},
      primaryClass={cs.CL},
      url={https://arxiv.org/abs/2305.14251}, 
}

@misc{zheng2023judging,
      title={Judging LLM-as-a-Judge with MT-Bench and Chatbot Arena}, 
      author={Lianmin Zheng and Wei-Lin Chiang and Ying Sheng and Siyuan Zhuang and Zhanghao Wu and Yonghao Zhuang and Zi Lin and Zhuohan Li and Dacheng Li and Eric P. Xing and Hao Zhang and Joseph E. Gonzalez and Ion Stoica},
      year={2023},
      eprint={2306.05685},
      archivePrefix={arXiv},
      primaryClass={cs.CL},
      url={https://arxiv.org/abs/2306.05685}, 
}

@misc{luo2023chatgpt,
      title={ChatGPT as a Factual Inconsistency Evaluator for Text Summarization}, 
      author={Zheheng Luo and Qianqian Xie and Sophia Ananiadou},
      year={2023},
      eprint={2303.15621},
      archivePrefix={arXiv},
      primaryClass={cs.CL},
      url={https://arxiv.org/abs/2303.15621}, 
}

@misc{liu2025suceareasoningintensiveretrievaladversarial,
      title={SUCEA: Reasoning-Intensive Retrieval for Adversarial Fact-checking through Claim Decomposition and Editing}, 
      author={Hongjun Liu and Yilun Zhao and Arman Cohan and Chen Zhao},
      year={2025},
      eprint={2506.04583},
      archivePrefix={arXiv},
      primaryClass={cs.CL},
      url={https://arxiv.org/abs/2506.04583}, 
}

@misc{park2023generativeagentsinteractivesimulacra,
      title={Generative Agents: Interactive Simulacra of Human Behavior}, 
      author={Joon Sung Park and Joseph C. O'Brien and Carrie J. Cai and Meredith Ringel Morris and Percy Liang and Michael S. Bernstein},
      year={2023},
      eprint={2304.03442},
      archivePrefix={arXiv},
      primaryClass={cs.HC},
      url={https://arxiv.org/abs/2304.03442}, 
}

@misc{zhong2023memorybankenhancinglargelanguage,
      title={MemoryBank: Enhancing Large Language Models with Long-Term Memory}, 
      author={Wanjun Zhong and Lianghong Guo and Qiqi Gao and He Ye and Yanlin Wang},
      year={2023},
      eprint={2305.10250},
      archivePrefix={arXiv},
      primaryClass={cs.CL},
      url={https://arxiv.org/abs/2305.10250}, 
}

@misc{shinn2023reflexionlanguageagentsverbal,
      title={Reflexion: Language Agents with Verbal Reinforcement Learning}, 
      author={Noah Shinn and Federico Cassano and Edward Berman and Ashwin Gopinath and Karthik Narasimhan and Shunyu Yao},
      year={2023},
      eprint={2303.11366},
      archivePrefix={arXiv},
      primaryClass={cs.AI},
      url={https://arxiv.org/abs/2303.11366}, 
}

@misc{zhao2024expelllmagentsexperiential,
      title={ExpeL: LLM Agents Are Experiential Learners}, 
      author={Andrew Zhao and Daniel Huang and Quentin Xu and Matthieu Lin and Yong-Jin Liu and Gao Huang},
      year={2024},
      eprint={2308.10144},
      archivePrefix={arXiv},
      primaryClass={cs.LG},
      url={https://arxiv.org/abs/2308.10144}, 
}

@misc{xu2025amemagenticmemoryllm,
      title={A-MEM: Agentic Memory for LLM Agents}, 
      author={Wujiang Xu and Zujie Liang and Kai Mei and Hang Gao and Juntao Tan and Yongfeng Zhang},
      year={2025},
      eprint={2502.12110},
      archivePrefix={arXiv},
      primaryClass={cs.CL},
      url={https://arxiv.org/abs/2502.12110}, 
}

@misc{zhang2024surveymemorymechanismlarge,
      title={A Survey on the Memory Mechanism of Large Language Model based Agents}, 
      author={Zeyu Zhang and Xiaohe Bo and Chen Ma and Rui Li and Xu Chen and Quanyu Dai and Jieming Zhu and Zhenhua Dong and Ji-Rong Wen},
      year={2024},
      eprint={2404.13501},
      archivePrefix={arXiv},
      primaryClass={cs.AI},
      url={https://arxiv.org/abs/2404.13501}, 
}

@article{laban-etal-2022-summac,
    title = "{S}umma{C}: Re-Visiting {NLI}-based Models for Inconsistency Detection in Summarization",
    author = "Laban, Philippe  and
      Schnabel, Tobias  and
      Bennett, Paul N.  and
      Hearst, Marti A.",
    editor = "Roark, Brian  and
      Nenkova, Ani",
    journal = "Transactions of the Association for Computational Linguistics",
    volume = "10",
    year = "2022",
    address = "Cambridge, MA",
    publisher = "MIT Press",
    url = "https://aclanthology.org/2022.tacl-1.10/",
    doi = "10.1162/tacl_a_00453",
    pages = "163--177"
}

@inproceedings{honovich-etal-2022-true-evaluating,
    title = "{TRUE}: Re-evaluating Factual Consistency Evaluation",
    author = "Honovich, Or  and
      Aharoni, Roee  and
      Herzig, Jonathan  and
      Taitelbaum, Hagai  and
      Kukliansy, Doron  and
      Cohen, Vered  and
      Scialom, Thomas  and
      Szpektor, Idan  and
      Hassidim, Avinatan  and
      Matias, Yossi",
    editor = "Carpuat, Marine  and
      de Marneffe, Marie-Catherine  and
      Meza Ruiz, Ivan Vladimir",
    booktitle = "Proceedings of the 2022 Conference of the North American Chapter of the Association for Computational Linguistics: Human Language Technologies",
    month = jul,
    year = "2022",
    address = "Seattle, United States",
    publisher = "Association for Computational Linguistics",
    url = "https://aclanthology.org/2022.naacl-main.287/",
    doi = "10.18653/v1/2022.naacl-main.287",
    pages = "3905--3920"
}

@article{rashkin-etal-2023-measuring,
    title = "Measuring Attribution in Natural Language Generation Models",
    author = "Rashkin, Hannah  and
      Nikolaev, Vitaly  and
      Lamm, Matthew  and
      Aroyo, Lora  and
      Collins, Michael  and
      Das, Dipanjan  and
      Petrov, Slav  and
      Tomar, Gaurav Singh  and
      Turc, Iulia  and
      Reitter, David",
    journal = "Computational Linguistics",
    volume = "49",
    number = "4",
    month = dec,
    year = "2023",
    address = "Cambridge, MA",
    publisher = "MIT Press",
    url = "https://aclanthology.org/2023.cl-4.2/",
    doi = "10.1162/coli_a_00486",
    pages = "777--840"
}

@inproceedings{gao-etal-2023-enabling,
    title = "Enabling Large Language Models to Generate Text with Citations",
    author = "Gao, Tianyu  and
      Yen, Howard  and
      Yu, Jiatong  and
      Chen, Danqi",
    editor = "Bouamor, Houda  and
      Pino, Juan  and
      Bali, Kalika",
    booktitle = "Proceedings of the 2023 Conference on Empirical Methods in Natural Language Processing",
    month = dec,
    year = "2023",
    address = "Singapore",
    publisher = "Association for Computational Linguistics",
    url = "https://aclanthology.org/2023.emnlp-main.398/",
    doi = "10.18653/v1/2023.emnlp-main.398",
    pages = "6465--6488"
}

@inproceedings{gao-etal-2023-rarr,
    title = "{RARR}: Researching and Revising What Language Models Say, Using Language Models",
    author = "Gao, Luyu  and
      Dai, Zhuyun  and
      Pasupat, Panupong  and
      Chen, Anthony  and
      Chaganty, Arun Tejasvi  and
      Fan, Yicheng  and
      Zhao, Vincent  and
      Lao, Ni  and
      Lee, Hongrae  and
      Juan, Da-Cheng  and
      Guu, Kelvin",
    editor = "Rogers, Anna  and
      Boyd-Graber, Jordan  and
      Okazaki, Naoaki",
    booktitle = "Proceedings of the 61st Annual Meeting of the Association for Computational Linguistics (Volume 1: Long Papers)",
    month = jul,
    year = "2023",
    address = "Toronto, Canada",
    publisher = "Association for Computational Linguistics",
    url = "https://aclanthology.org/2023.acl-long.910/",
    doi = "10.18653/v1/2023.acl-long.910",
    pages = "16477--16508"
}

@inproceedings{maharana-etal-2024-evaluating,
    title = "Evaluating Very Long-Term Conversational Memory of {LLM} Agents",
    author = "Maharana, Adyasha  and
      Lee, Dong-Ho  and
      Tulyakov, Sergey  and
      Bansal, Mohit  and
      Barbieri, Francesco  and
      Fang, Yuwei",
    editor = "Ku, Lun-Wei  and
      Martins, Andre  and
      Srikumar, Vivek",
    booktitle = "Proceedings of the 62nd Annual Meeting of the Association for Computational Linguistics (Volume 1: Long Papers)",
    month = aug,
    year = "2024",
    address = "Bangkok, Thailand",
    publisher = "Association for Computational Linguistics",
    url = "https://aclanthology.org/2024.acl-long.747/",
    doi = "10.18653/v1/2024.acl-long.747",
    pages = "13851--13870"
}

@misc{chen2026halumemevaluatinghallucinationsmemory,
      title={HaluMem: Evaluating Hallucinations in Memory Systems of Agents}, 
      author={Ding Chen and Simin Niu and Kehang Li and Peng Liu and Xiangping Zheng and Bo Tang and Xinchi Li and Feiyu Xiong and Zhiyu Li},
      year={2026},
      eprint={2511.03506},
      archivePrefix={arXiv},
      primaryClass={cs.CL},
      url={https://arxiv.org/abs/2511.03506}, 
}
